%% file: colm2026_conference.tex
\documentclass{article} 

\makeatletter
\def\bibroot{}
\IfFileExists{colm2026_conference.sty}{}{\def\bibroot{Template-2026/}}
\providecommand\input@path{}
\edef\input@path{{./}{Template-2026/}}
\makeatother

\usepackage[final]{colm2026_conference}

\usepackage{amsmath}
\usepackage{microtype}
\usepackage{hyperref}
\usepackage{cleveref}
\usepackage{url}
\usepackage{booktabs}
\usepackage{xspace}
\usepackage{tikz}
 \usepackage{subcaption}
 \usepackage[table]{xcolor}
\usepackage[dvipsnames]{xcolor}
\usepackage{multirow} 
 \usepackage{siunitx}
 \usepackage{algorithm} 
 \usepackage{algorithmic} 
 \usepackage{wrapfig}
 \usepackage{enumitem}

\usetikzlibrary{positioning, arrows.meta, shapes.geometric, calc, backgrounds, fit}
\usetikzlibrary{shapes.geometric, arrows.meta, positioning, calc, shadows, fit, backgrounds, decorations.pathreplacing}

\usepackage{lineno}

\definecolor{ourscolor}{gray}{0.92}
\definecolor{darkblue}{rgb}{0, 0, 0.5}
\hypersetup{colorlinks=true, citecolor=darkblue, linkcolor=darkblue, urlcolor=darkblue}

\usepackage[compact]{titlesec}
\titlespacing{\section}{0pt}{*1}{*0}
\titlespacing{\subsection}{0pt}{*1}{*0}

\usepackage[subtle, mathdisplays=tight, charwidths=tight, leading=normal]{savetrees}

\graphicspath{{./}{Template-2026/}}

\title{Thought-Level Beam Search for Reasoning}

\author{Lijie Yang \\
Princeton University \\
\texttt{ly3223@princeton.edu} \\
\And
Hongyin Luo \\
MIT CSAIL \\
\texttt{hyluo@mit.edu} \\
\And
Jiawei Zhao \\
Meta AI \\
\texttt{jwzhao@meta.com} \\
\AND
Tri Dao\thanks{Equal advising.} \\
Princeton University \\
\texttt{tridao@princeton.edu} \\
\And
Ravi Netravali\footnotemark[\value{footnote}] \\
Princeton University \\
\texttt{rnetravali@cs.princeton.edu} \\
}

\newcommand{\sysname}{Gambit\xspace}
\newcommand{\scorername}{SeqScorer\xspace}
\newcommand{\TODO}[1]{{\textcolor{blue}{TODO: #1}}}

\newcommand{\green}[1]{\textcolor{ForestGreen}{#1}}
\newcommand{\TD}[1]{{\textcolor{cyan}{TD: #1}}}
\newcommand{\rn}[1]{{\textcolor{red}{RN: #1}}}

\definecolor{phaseGreen}{HTML}{25664A}
\definecolor{phaseOrange}{HTML}{D4770B}
\definecolor{phaseBlue}{HTML}{4C78A8}
\newcommand{\phaselabel}[2]{%
  \colorbox{#1}{\strut\textcolor{white}{\small\textsc{#2}}}%
}

\begin{document}

\ifcolmsubmission
\linenumbers
\fi

\maketitle

\begin{abstract}
Test-time compute scaling is a primary driver of performance in large reasoning models (LRMs), but extreme inefficiency bounds current approaches, shifting the critical question from \emph{how much} compute to spend, to \emph{where} to allocate it. We formalize test-time reasoning as a constrained compute allocation problem over partial trajectories. Under a fixed hardware budget, existing paradigms fail to actively allocate the compute to the most promising partial progress: traditional parallel sampling treats traces independently and induces severe memory bottlenecks, while subtractive pruning starves hardware and fails to actively and sufficiently shift the output distribution. To overcome this dichotomy, we introduce \sysname{}, an inference algorithm that executes \emph{thought-level beam search}. By periodically pruning unpromising trajectories and immediately branching from high-quality prefixes, \sysname{} dynamically concentrates compute onto the most promising reasoning traces via a light-weight scorer probing hidden states while maintaining continuous high hardware utilization. Extensive evaluations across multiple models and benchmarks demonstrate that \sysname{} strictly dominates existing baselines. Under identical hardware constraints, our method yields up to a +6.7\% absolute accuracy gain on HMMT-24 and +3.3\% on AIME-25 over pruning baselines, delivers $>2\times$ higher throughput on trace completion, and reduces total token consumption by up to 68.5\% relative to standard parallel sampling
\footnote{Code is available at \href{https://github.com/Dao-AILab/Gambit.git}{https://github.com/Dao-AILab/Gambit.git}.}.
\end{abstract}

\input{sections/introduction}

\input{sections/related_work}
\input{sections/motivation}
\input{sections/method}
\input{sections/experiments}

\input{sections/ablation}
\input{sections/conclusion}

\newpage
\section*{Acknowledgments}
We are grateful to the compute support at Dao AI Lab and insightful discussions with Wentao Guo, Rui Pan, Yizheng Zhang, Alan Zhu, and Han Guo. 

\bibliography{\bibroot colm2026_conference}
\bibliographystyle{\bibroot colm2026_conference}

\newpage
\appendix
\section{Appendix}
\label{sec:appendix}
\input{sections/appendix}

\end{document}

%% file: sections/introduction.tex
\vspace{-1em}
\section{Introduction}
\label{sec:intro}
Large reasoning models (LRMs) have recently demonstrated impressive capabilities on challenging mathematical and scientific tasks by generating long chains of thought during inference \citep{anthropic2025claude4, openai2025gpt5, openai2025gptoss, deepseekai2025deepseekr1incentivizingreasoningcapability, wei2023chainofthoughtpromptingelicitsreasoning, qwen3technicalreport}.
A key driver of these accuracy gains is \textbf{test-time compute scaling}, or allocating additional compute during inference, typically through parallel sampling of reasoning traces~\citep{muennighoff2025s1simpletesttimescaling, kojima2023largelanguagemodelszeroshot}.
Self-consistency, which aggregates answers from multiple independently sampled traces, is the canonical implementation of this strategy \citep{wang2023selfconsistencyimproveschainthought}.

However, the current paradigm is reaching diminishing returns due to extreme inefficiency. As shown in \Cref{fig:hero_topology}, parallel sampling requires generating hundreds of long reasoning traces, the vast majority of which can lead to incorrect answers.
Specifically, one NVIDIA B300, the state-of-the-art hardware, needs several hours to complete 512 traces for a single AIME-2025 problem on vLLM ~\citep{kwon2023efficientmemorymanagementlarge} with Qwen3-8B model ~\citep{qwen3technicalreport} while still failing on the hardest instances.
As a result, it becomes necessary to ensure that the scaled inference can effectively lead to performance improvements. In other words, the central question for the field has shifted from \emph{\textbf{how much}} compute to spend to \emph{\textbf{where}} to allocate it. 

\input{figures/intro/main_figure}

At its core, test-time reasoning can be viewed as a problem of allocating compute over a population of partial trajectories under strict hardware constraints.
Making this allocation effective requires resolving three tightly coupled challenges:(1) \emph{checkpointing partial trajectories worth continuing}, (2) \emph{allocating additional compute to amplify them}, and (3) \emph{doing so while keeping the hardware fully utilized}.

\paragraph{Checkpointing promising partial trajectories.}
Reasoning is inherently fragile, yet successful and failed trajectories often share rigorous intermediate prefixes before diverging~\citep{lightman2023letsverifystepstep, dziri2023faithfatelimitstransformers, song2026largelanguagemodelreasoning, uesato2022solvingmathwordproblems}. Capturing these states during generation is therefore critical. Existing evaluators—from costly Process Reward Models~\citep{wang2024mathshepherdverifyreinforcellms, xie2024montecarlotreesearch, yao2023treethoughtsdeliberateproblem} to lightweight internal signals~\citep{fu2025deepthinkconfidence, kadavath2022languagemodelsmostlyknow, lin2022teachingmodelsexpressuncertainty}—can identify promising steps, but without checkpointing, a single downstream error discards all prior valid progress, wasting previous compute. 
Naive reliance on evaluators is also brittle-early reasoning is incomplete and yields noisy, unreliable signals~\citep{hao2023reasoninglanguagemodelplanning}. An effective strategy must therefore combine efficient evaluation with principled control over \emph{when} a prefix is reliable enough to checkpoint.


\paragraph{Allocating compute to amplify promising trajectories.}
Yet, successfully checkpointing a reliable prefix is only the first step. On challenging problems, correct solutions are exceedingly rare and easily overwhelmed by plausible but incorrect reasoning paths~\citep{lee2025evaluatingstepbystepreasoningtraces, fu2025deepthinkconfidence}. Consequently, purely subtractive strategies—which attempt to improve efficiency and accuracy by merely terminating low-quality traces—are fundamentally insufficient~\citep{liang2026hiddenstatesearlysignals, hong-etal-2025-slim}. That is because the underlying output distribution is largely left unchanged. To overcome this, freed compute must be \emph{actively reallocated} to expand high-quality prefixes into more continuations, fundamentally shifting the output distribution toward the correct answer.

\paragraph{Sustaining hardware utilization.}
Finally, active reallocation must respect strict GPU memory constraints. Unmanaged parallel sampling saturates memory early, inducing severe queueing delays in modern serving engines~\citep{kwon2023efficientmemorymanagementlarge, zheng2024sglangefficientexecutionstructured}. Conversely, pruning-based methods~\citep{liang2026hiddenstatesearlysignals, fu2025deepthinkconfidence, hong-etal-2025-slim} alleviate memory pressure but induce hardware starvation: terminated traces are discarded without replacement, leaving GPU resources increasingly idle. An optimal approach should balance these extremes, maintaining a constant level of active computation to ensure sustained hardware utilization throughout decoding.

To address these limitations, we propose \textbf{\sysname}, an inference algorithm that formulates test-time reasoning as a \emph{thought-level beam search} over partial trajectories.
Our design mirrors the three challenges above.
First, we evaluate and checkpoint promising prefixes using efficient, model-internal signals—either off-the-shelf hidden-state probes adapted from STEP~\citep{liang2026hiddenstatesearlysignals} or a custom sequence scorer. To overcome the inherent early-stage noise of these signals, \sysname{} introduces a strict \emph{warmup threshold}. This algorithmic constraint delays all scoring and checkpointing until a sufficient reasoning depth is reached, ensuring the system only checkpoints logically stable foundations.
Second, we amplify these prefixes through a periodic \textbf{beam search} procedure that performs synchronized prune-and-branch operations, reallocating compute toward high-quality trajectories.
Third, we enforce a \emph{zero-sum allocation policy} that maintains a constant-size pool of active traces, ensuring sustained hardware utilization throughout generation. Furthermore, to balance exploration and exploitation and prevent greedy collapse, \sysname{} utilizes a decoupled memory management scheme to avoid aggressively branching over a single top trajectory.
Together, these components transform test-time scaling from passive sampling into an active, system-aware search process. 

To validate our approach, we deploy \sysname{} system atop vLLM and evaluate across a suite of challenging reasoning benchmarks (AIME, HMMT, and GPQA-Diamond). Across varying model architectures, active thought-level search strictly dominates both static and subtractive baselines. Under identical hardware constraints, \sysname{} improves accuracy by \textbf{+6.7\%} on HMMT-24 and \textbf{+3.3\%} on AIME-25 over aggressive pruning methods, while reducing total token consumption by up to \textbf{68.5\%} relative to standard parallel sampling. Furthermore, \sysname{} sustains continuous hardware saturation, delivering over \textbf{$2\times$} higher productive trace throughput with $<1\%$ system overhead.

%% file: figures/intro/main_figure.tex
\begin{figure}[!ht]
    \centering
    \includegraphics[width=0.85\textwidth]{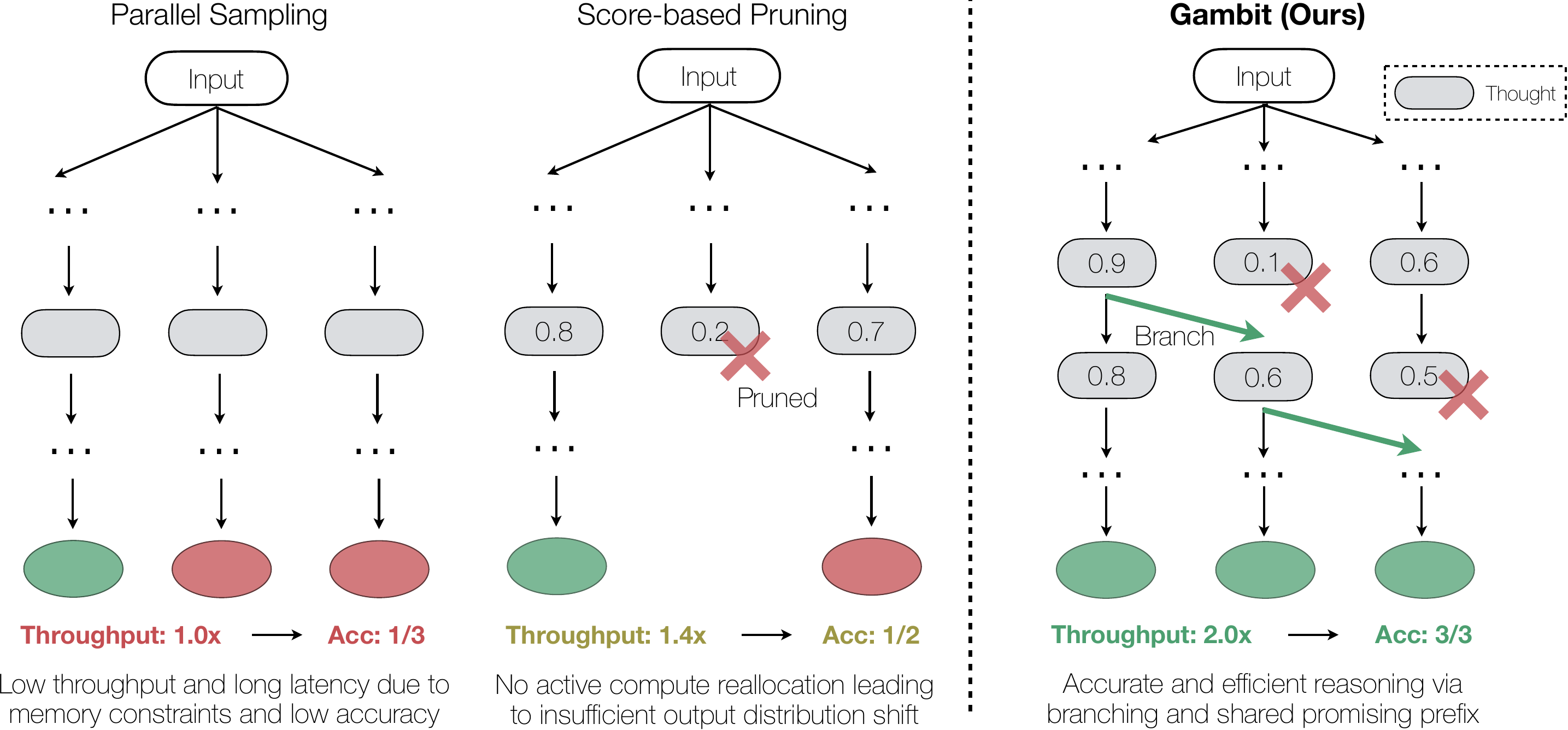}
    \caption{
        \textbf{Reasoning topologies for test-time compute allocation.} 
        \textbf{(Left)} Parallel sampling treats traces as independent trials, leading to wrong answers in majority.
        \textbf{(Center)} Score-based pruning terminates unpromising paths but leaves the freed capacity idle, failing to actively shift the sampling distribution.
        \textbf{(Right)} \sysname{} actively reallocates compute: when low-scoring traces are pruned, it immediately branches from high-quality prefixes to amplify correct reasoning and saturate the compute budget.
    }
    \label{fig:hero_topology}
\end{figure}

%% file: sections/related_work.tex
\vspace{-1em}
\section{Related Work}
\label{sec:related}

\paragraph{Test-Time Compute Scaling.}
Scaling compute at inference time is a primary driver of modern reasoning performance \citep{openai2025gptoss, deepseekai2025deepseekr1incentivizingreasoningcapability, muennighoff2025s1simpletesttimescaling, gemmateam2025gemma3technicalreport}. The popular paradigm, Self-Consistency (SC) \citep{wang2023selfconsistencyimproveschainthought}, improves accuracy by aggregating traces via majority voting. However, SC scales highly inefficiently: because trajectories are sampled entirely independently, the model repeatedly allocates compute to the redundant exploration of degenerate pathways \citep{song2026largelanguagemodelreasoning}. 

\paragraph{Inference-Time Search.}
To improve sample efficacy, methods such as Tree-of-Thoughts (ToT) \citep{yao2023treethoughtsdeliberateproblem} and Monte Carlo Tree Search (MCTS) \citep{xie2024montecarlotreesearch} frame reasoning as a structured search guided by value function or process reward models (PRMs) \citep{lightman2023letsverifystepstep, wang2024mathshepherdverifyreinforcellms}. While MCTS balances exploration and exploitation, its asymmetric expansion and asynchronous rollouts clash with the synchronous, large-batch nature of modern LLM inference. \sysname{} bridges this gap by reformulating search as a hardware-constrained, synchronous beam allocation by maintaining a constant-size active pool via zero-sum prune-and-branch operations.

\paragraph{Efficient Serving and Pruning-based Approaches.}
Large-batch, long-context reasoning serving imposes severe KV-cache pressure \citep{kwon2023efficientmemorymanagementlarge, zheng2024sglangefficientexecutionstructured}. To mitigate this, recent works propose subtractive pruning: DeepConf \citep{fu2025deepthinkconfidence} and STEP \citep{liang2026hiddenstatesearlysignals} early-terminate unpromising traces using internal model signals, while Slim-SC \citep{hong-etal-2025-slim} deduplicates via similarity. However, these purely subtractive methods terminate traces without reallocating the freed compute, failing to sufficiently shift the underlying generation distribution. \sysname{} resolves this by active compute reallocation, dynamically amplifying the presence of correct traces in the ensemble while eliminating the hardware starvation inherent to pruning.

%% file: sections/motivation.tex
\vspace{-1em}
\section{Motivation}
\label{sec:motivation}
\input{figures/motivation/branching_motivation}
As we discuss in \Cref{sec:intro}, an effective test-time strategy in reasoning must (i) allocate compute toward promising partial trajectories and (ii) maintain high hardware utilization. 
We now provide two empirical observations motivating these requirements.

\paragraph{High-quality prefixes manifest enhanced compute efficacy.} 
Recent work shows that successful and failed trajectories often exhibit promising high-quality prefixes well before completion~\citep{uesato2022solvingmathwordproblems, wang2024mathshepherdverifyreinforcellms}, indicating that useful computation is disproportionately concentrated in early intermediate states. We quantify this in \Cref{fig:prefix_value}: pausing 64 parallel traces midway, ranking them with a lightweight hidden-state scorer adapted from STEP~\citep{liang2026hiddenstatesearlysignals}, and branching exclusively from the highest-scoring prefix yields dramatic accuracy gains while consuming roughly half the token budget. We observe similarly large lifts on other hard problems (e.g., $1.6\% \to 39.1\%$ on AIME-25 Q12). While these results highlight the extreme inefficiency of uniform parallel sampling, naively over-committing to a single prefix easily leads to confident but incorrect trajectories~\citep{fu2025deepthinkconfidence}. This necessitates \emph{controlled reallocation}---maintaining a beam of candidates while progressively concentrating compute on the most promising subset.

\input{figures/motivation/system_motivation}

\paragraph{Reasoning workloads underutilize hardware.}
Efficiently serving long reasoning traces in large batches requires balancing concurrency with KV-cache capacity.
Standard parallel sampling launches many concurrent traces, but long contexts quickly saturate the KV cache. As illustrated in \Cref{fig:utilization}, this forces inference engines such as vLLM \citep{kwon2023efficientmemorymanagementlarge} or SGLang \citep{zheng2024sglangefficientexecutionstructured} to severely queue requests, heavily inflating end-to-end latency.
Pruning-based approaches alleviate memory pressure \citep{fu2025deepthinkconfidence, liang2026hiddenstatesearlysignals}, but introduce the opposite problem: hardware starvation. Because pruned traces are permanently discarded and not replaced, concurrency steadily declines during generation. As shown in \Cref{fig:utilization}, this leaves GPU resources increasingly idle for the remainder of the generation window.

%% file: figures/motivation/branching_motivation.tex
\begin{figure}[!h]
    \centering
    \includegraphics[width=0.8\textwidth]{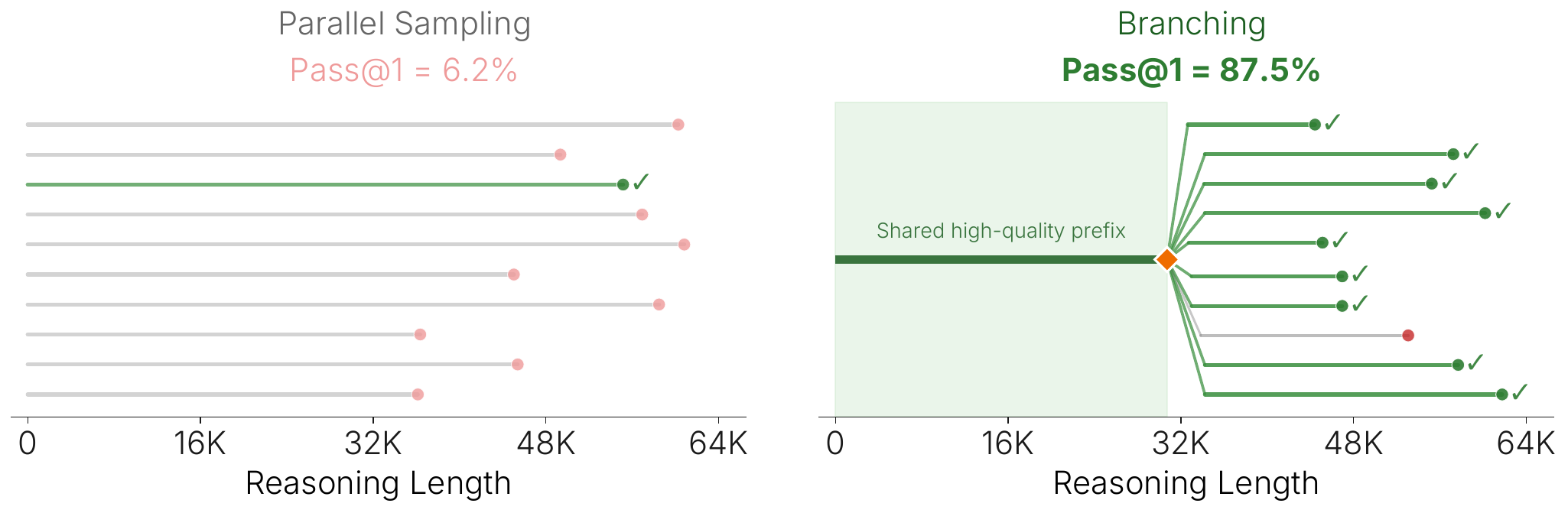}
    \caption{
        Branching from a high-quality prefix versus independent sampling on a hard AIME~2025 problem (Q27).
        Branching 64 continuations from the top-ranked prefix achieves 87.5\% pass@1 ($14\times$ over 6.2\% baseline) while sharing the parent's KV-cache (green region) halves memory consumption.
    }
    \label{fig:prefix_value}
\end{figure}

%% file: figures/motivation/system_motivation.tex
\begin{figure}[!h]
    \centering
    \includegraphics[width=0.8\linewidth]{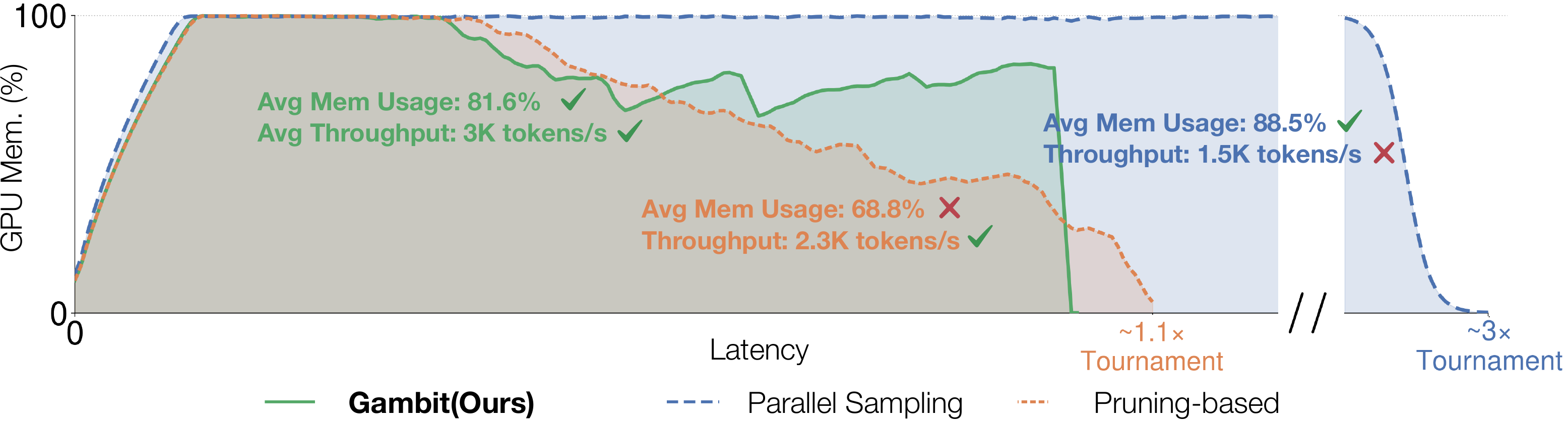}
    \caption{
        GPU memory utilization and latency profiles during reasoning with batch\_size=256 traces on HMMT-25 Q7.
        Parallel sampling exhausts KV-cache capacity, inflating latency by ${\sim}3\times$.
        Pruning-only STEP~\citep{liang2026hiddenstatesearlysignals} eliminates queueing but steadily loses concurrency.
        \sysname{} maintains high utilization throughout at ${\sim}1.1\times$ latency.
    }
    \label{fig:utilization}
\end{figure}

%% file: sections/method.tex
\vspace{-0.5em}
\section{Method: \sysname{}}
\label{sec:method}

Motivated by the observations in \Cref{sec:motivation}, we formalize test-time reasoning as a \emph{constrained compute-allocation problem} over a dynamic population of partial reasoning trajectories. Given a problem prompt $\mathcal{P}$, the objective is to discover a thought-level allocation policy $\pi$ that maximizes the probability of deriving the ground-truth solution $y^*$, subject to a strict inference-time hardware budget:
\vspace{-0.1em}
\begin{equation}
\max_{\pi} \;\; \mathbb{P}\big( \mathcal{A}_\pi(\mathcal{P}) = y^* \mid \mathcal{P} \big)
\qquad \text{s.t.} \qquad
\Omega(\pi) \le B,
\label{eq:objective}
\end{equation}
where $\mathcal{A}_\pi(\mathcal{P})$ denotes the final aggregated answer produced by policy $\pi$, and $\Omega(\pi)$ represents the peak computational and memory cost incurred during execution. In the context of autoregressive large language models, this hardware budget $B$ manifests primarily as a hard physical constraint on the maximum number of concurrently active traces (preventing compute starvation) and the total KV-cache memory footprint.

A key empirical premise, established in \Cref{sec:motivation}, is that hidden-state-based scoring functions can identify promising reasoning prefixes midway through generation, and that continuing from such prefixes yields a higher probability of producing a correct answer. This implies that uniform allocation of compute across independent traces is suboptimal.
Inspired by beam search ~\citep{lowerre1976harpy, sutskever2014sequencesequencelearningneural}, we therefore instantiate inference as a \emph{thought-level beam search}. Unlike classical token-level beam search—which optimizes sequence likelihood—our objective is to allocate limited compute toward prefixes most likely to yield a correct solution under a strict capacity constraint. Concretely, we operate at the granularity of reasoning steps, use a scoring function as a proxy for continuation value, and maintain a fixed-size active pool throughout execution.

To realize this policy under real hardware constraints, we introduce a fixed-budget algorithm that performs periodic, zero-sum reallocation of compute across traces while maintaining a constant memory footprint (\Cref{sec:algorithm}). We further design a decoupled memory management scheme that separates logical search decisions from physical execution state, preventing pathological accuracy collapse under memory pressure (\Cref{sec:decoupled_sys_design}).

\subsection{Thought-Level Beam Search}
\label{sec:algorithm}
\input{figures/method/pipeline}

We now describe our solution to the constrained compute-allocation problem defined in \Cref{eq:objective}. The full algorithm and visualization are in \Cref{alg:sysname} and \Cref{fig:pipeline}. The central challenge is to allocate limited decoding capacity across a population of partial reasoning trajectories so as to maximize the probability of producing a correct final answer.

\begin{algorithm}[ht!]
\fontsize{7.5pt}{9pt}\selectfont
\caption{Thought-Level Beam Search in \sysname{})}
\label{alg:sysname}
\textbf{Input:} Problem prompt $\mathcal{P}$,\; capacity $C$,\; swap size $K$,\; check interval $\Delta$,\; warmup $w$,\; scorer $f_\theta$ \\
\textbf{Output:} Score-weighted majority-vote answer $a^*$
\vspace{0.1em}
\hrule
\vspace{0.1em}
\begin{minipage}[t]{0.48\textwidth}
\begin{algorithmic}[1]
\STATE \textbf{Main Loop:}
\STATE $\mathcal{S} \gets \{C \text{ initial traces from } \mathcal{P}\}$
  \hfill{\color{gray}\itshape\footnotesize$\triangleright$ Scheduler view}
\STATE $\mathcal{T}_{\text{act}} \gets \mathcal{S}$
  \hfill{\color{gray}\itshape\footnotesize$\triangleright$ Tree view}
\STATE $\mathcal{F} \gets \emptyset$
  \hfill{\color{gray}\itshape\footnotesize$\triangleright$ Completed traces}
\WHILE{$|\mathcal{S}| > 0$ \textbf{and} $|\mathcal{F}| < C$}
    \STATE \phaselabel{phaseGreen}{Generate}
      \; Advance one thought $\forall\, \tau \in \mathcal{S}$
    \STATE For each $\tau$ completing step $s_n$:
    \STATE \quad $\bar{s}_\tau \gets \bar{s}_\tau + \frac{1}{n}\bigl(f_\theta(\mathbf{h}_n) - \bar{s}_\tau\bigr)$
      \hfill{\color{gray}\itshape\footnotesize$\triangleright$ Cumulative score}
    \STATE \phaselabel{phaseOrange}{Schedule}
      \; \textit{On GPU memory pressure}
    \IF{KV-cache is saturated}
        \STATE $\tau_{\text{vic}} \gets \arg\min_{\tau \in \mathcal{S}} \bar{s}_\tau$
        \STATE $\mathcal{S} \gets \mathcal{S} \setminus \{\tau_{\text{vic}}\}$
          \hfill{\color{gray}\itshape\footnotesize$\triangleright$ Evict as ghost trace}
    \ENDIF
    \STATE \phaselabel{phaseBlue}{Tournament}
      \; \textit{Every $\Delta$ thoughts}
    \IF{tournament condition met}
        \STATE $\mathcal{T}_{\text{act}},\, \mathcal{S} \gets \textsc{Tournament}(\mathcal{T}_{\text{act}},\, \mathcal{S},\, C,\, K)$
    \ENDIF
    \STATE Move completed traces to $\mathcal{F}$
\ENDWHILE
\end{algorithmic}
\end{minipage}%
\hfill
\vrule
\hfill
\begin{minipage}[t]{0.48\textwidth}
\begin{algorithmic}[1]
\STATE \textbf{Function} \textsc{Tournament}$(\mathcal{T}_{\text{act}},\, \mathcal{S},\, C,\, K)$:
\STATE $N \gets |\mathcal{T}_{\text{act}}|$
\STATE Rank $\tau \in \mathcal{T}_{\text{act}}$ by $\bar{s}_\tau$ descending
\STATE $\mathcal{B} \gets \{\tau \in \mathcal{T}_{\text{act}} : \text{GenLen}(\tau) \geq w\}$
  \hfill{\color{gray}\itshape\footnotesize$\triangleright$ Eligible parents}
\IF{$N < C$}
    \STATE $m \gets \min(C - N,\, |\mathcal{B}|)$
      \hfill{\color{gray}\itshape\footnotesize$\triangleright$ \textbf{Case 1:} Under-capacity}
    \STATE $\mathcal{C}_{\text{new}} \gets \mathrm{Branch}(\text{Top-}m \text{ from } \mathcal{B})$
\ELSIF{$N = C$}
    \STATE $\mathcal{V} \gets \text{Bottom-}K \text{ from } \mathcal{T}_{\text{act}}$
      \hfill{\color{gray}\itshape\footnotesize$\triangleright$ \textbf{Case 2:} At capacity}
    \STATE $\mathcal{T}_{\text{act}} \gets \mathcal{T}_{\text{act}} \setminus \mathcal{V}$;\;
           $\mathcal{S} \gets \mathcal{S} \setminus \mathcal{V}$
      \hfill{\color{gray}\itshape\footnotesize$\triangleright$ Prune bottom-k}
    \STATE $\mathcal{C}_{\text{new}} \gets \mathrm{Branch}(\text{Top-}K \text{ from } \mathcal{B})$
      \hfill{\color{gray}\itshape\footnotesize$\triangleright$ Branch top-k}
\ENDIF
\STATE $\mathcal{T}_{\text{act}} \gets \mathcal{T}_{\text{act}} \cup \mathcal{C}_{\text{new}}$
\STATE $\mathcal{S} \gets \mathcal{S} \cup \mathcal{C}_{\text{new}}$
\RETURN $\mathcal{T}_{\text{act}},\, \mathcal{S}$
\end{algorithmic}
\vspace{0.8em}
{\color{gray}\small
\textbf{Notation:} $\mathbf{h}_n$ = last-layer hidden state at step $n$;\; $\mathrm{Branch}(\tau)$ = spawn child from $\tau$'s prefix via KV-cache sharing;\; $w$ = warmup threshold (min steps before eligible for branching).
}
\end{minipage}
\vspace{0.1em}
\hrule
\vspace{0.1em}
\textbf{return} $a^* = \arg\max_{a}\;\sum_{\tau \in \mathcal{F}:\,\mathrm{ans}(\tau)=a} \bar{s}_\tau$
\end{algorithm}
\sysname{} performs a zero-sum reallocation of compute over a fixed-capacity pool of reasoning traces. At each round, the algorithm ranks all active trajectories using a scoring function, removes the lowest-scoring traces to free capacity, and immediately reallocates that capacity by branching from the highest-scoring prefixes. 

More specifically, each reasoning trajectory is modeled as a sequence of discrete steps (``thoughts'') separated by ``\texttt{\textbackslash n\textbackslash n}''~\citep{pan2025specreason}, denoted by $\tau = (s_1, s_2, \ldots, s_n)$ and $C$ be the operator-specified hardware \emph{capacity} (the maximum number of concurrent traces permitted by the system). \sysname{} operates in rounds triggered every $\Delta$ steps across all running traces (the \emph{check interval}).
At each tournament round, let $\mathcal{A}$ denote the set of currently active traces with $|\mathcal{A}| = N$. Let $\mathcal{B} \subseteq \mathcal{A}$ be the \emph{eligible branching candidates}—traces that have generated at least $w$ tokens since their own creation, a constraint that prevents cascading branch explosions from immature children.
Traces in $\mathcal{A}$ are ranked by their average score $\bar{s}_i = \frac{1}{n}\sum_{j=1}^{n} f_\theta(\mathbf{h}_{i,j})$, where $f_\theta$ can be any scorer function that takes $\mathbf{h}_{i,j}$, the last-layer hidden state at the boundary of step $j$ in trace $\tau_i$, as input. In the evaluation, \sysname{} uses the light-weight 2-Layer MLP scorer from STEP ~\citep{liang2026hiddenstatesearlysignals} or custom sequence scorer trained in \Cref{sec:scorer}.

It then forms an ordered list $\tau_{(1)}, \ldots, \tau_{(n)}$ with $\bar{s}_{(1)} \geq \cdots \geq \bar{s}_{(n)}$ and two routing cases arise:

\textbf{Case 1 — Under-capacity} ($N < C$): The pool has $C - N$ vacant slots. \sysname{} selects the top $\min(C - N, |\mathcal{B}|)$ eligible traces and \emph{branches} each one, filling the pool back to capacity:
\begin{equation}
    \text{Branch}\!\left(\tau_{(k)}\right) \;\text{ for } k = 1, \ldots, \min(C-N,\, |\mathcal{B}|). \label{eq:fill}
\end{equation}

\textbf{Case 2 — At capacity} ($N = C$): \sysname{} executes a rank-based \emph{swap} of size $K$:
\begin{align}
    &\text{Prune}\!\left(\tau_{(N-k+1)}\right);\  \text{Branch}\!\left(\tau_{(k)}\right) \;\text{ for } k = 1, \ldots, K, \label{eq:prune}
\end{align}
Pruning immediately terminates the $K$ lowest-scoring traces and frees their non-shared KV-cache blocks. Branching spawns $K$ new child requests from the $K$ highest-scoring prefixes. The child inherits the parent's KV-cache via prefix caching with minimal overhead. A temperature multiplier can be applied to the child to promote diversity. Because pruning and branching are paired identically, the total active count remains exactly $C$, enforcing a strict zero-sum memory invariant.

\subsection{Systems Integration: Decoupled Memory Management}
\label{sec:decoupled_sys_design}
A critical failure mode arises when logical search decisions are tightly coupled to physical execution state. Under memory pressure, the inference engine often preempts or, in our setting, prunes running traces to satisfy capacity constraints. If the search algorithm directly observes these evictions, the effective number of active traces drops below capacity, triggering repeated under-capacity branching on single Top-1/2 trace. This induces a pathological feedback loop in which compute is repeatedly concentrated on a small set of high-scoring prefixes, collapsing the search distribution and degrading solution quality.

To prevent this, \sysname{} introduces a \textbf{Decoupled View Architecture}, referred as Scheduler/Tree view in \Cref{alg:sysname}:
\begin{itemize}[nosep, leftmargin=1.5em]
    \item The Scheduler View (Memory Manager): Tracks only traces currently holding physical KV-cache blocks. It opportunistically evicts the lowest-ranked running trace when memory approaches saturation to prevent pushing traces to waiting queue.
    \item \textbf{The Tree View (Beam Search):} Tracks the logical topology of all active traces. Crucially, a trace evicted by the scheduler becomes a ``ghost trace''---it generates no new tokens and holds no physical memory, but its \texttt{is\_active} flag remains \texttt{True} in the logical tree.
\end{itemize}
By decoupling these views, \sysname{} executes its capacity checks against the \emph{Tree View} rather than the \emph{Scheduler View}. When the scheduler drops a trace, the logical count remains $N = C$. 
Consequently, the tournament correctly perceives the system as being at capacity and executes a balanced Case 2 swap (pruning the bottom-$K$ and branching the top-$K$), completely bypassing the greedy Case 1 fallback. Ghost traces simply retain their logical position until they naturally fall into the bottom-$K$ and are permanently pruned. This ensures that system-level scheduling does not distort the intended allocation policy over trajectories.

\paragraph{Termination \& Aggregation.} 
Because ghost traces remain logically active but generate no tokens, \sysname{} monitors the Scheduler View to determine termination. When the physical running count reaches zero, or when $C$ traces have successfully completed, generation stops. Each finished trace $\tau_i$ contributes its extracted answer $a_i$ with weight $\bar{s}_i$ to a score-weighted majority vote $a^* = \operatorname{arg\,max}_{a} \sum_{i:\, a_i = a} \bar{s}_i$.

%% file: figures/method/pipeline.tex
\begin{figure}[!ht]
\centering
\includegraphics[width=0.9\textwidth]{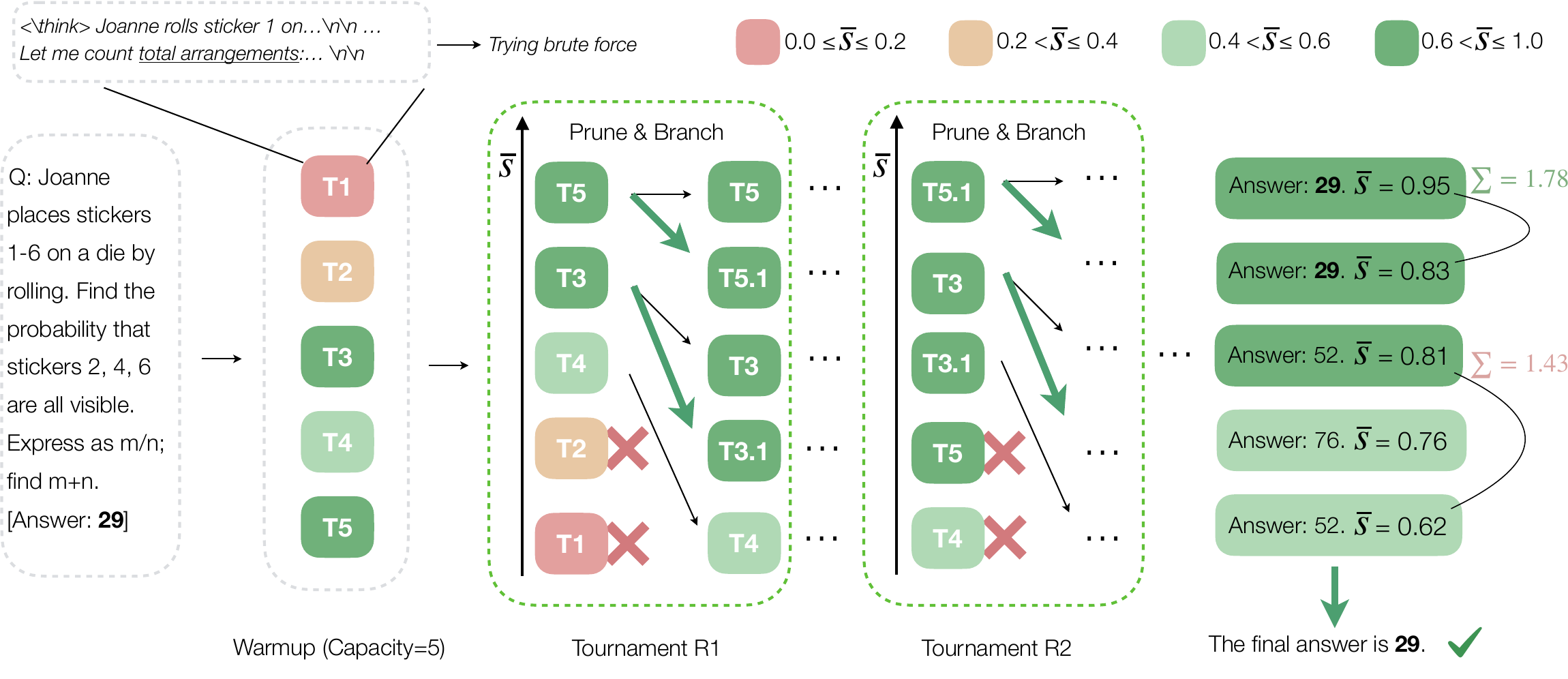}
\caption{
\textbf{End-to-end pipeline of \sysname{} on an AIME problem (capacity $C{=}5$, swap size $K{=}2$).}
During warmup, $C{=}5$ parallel traces are evaluated (color intensity denotes running score $\bar{s} \in [0,1]$).
Every $\Delta$ steps, a tournament ranks the active traces: the $K{=}2$ lowest-scoring traces are pruned ({\color{red}$\boldsymbol{\times}$}) and replaced by new branches spawned from the highest-scoring prefixes via prefix caching (green arrows). This zero-sum reallocation maintains exactly $C$ active traces throughout generation.
Upon completion, answers are aggregated via a score-weighted majority vote, correctly selecting answer 29 ($\sum \bar{s} = 1.78$ vs. $1.43$).
}
\label{fig:pipeline}
\end{figure}

%% file: sections/experiments.tex
\vspace{-1em}
\section{Experiments}
\label{sec:experiments}
Our empirical evaluation is designed to answer two primary research questions: 
(\emph{i}) Does \sysname{} strictly improve downstream reasoning accuracy over subtractive pruning and unmanaged sampling under the same hardware constraint? 
(\emph{ii}) How does active compute reallocation impact absolute token consumption and latency compared to those strategies?

\subsection{Experimental Setup}
\paragraph{Models \& Benchmarks.}
We evaluate across three open-weight large reasoning models of varying architectures and scales: Qwen3-4B-Thinking-2507~\citep{qwen3technicalreport}, DeepSeek-R1-0528-Qwen3-8B~\citep{qwen3technicalreport}, and Phi-4-reasoning-plus-14B~\citep{abdin2025phi4reasoningtechnicalreport}. 
This diversity allows us to assess whether score-guided search generalizes across model families.
We measure accuracy on four competition-grade mathematical reasoning benchmarks: AIME 2025, AIME 2026 ~\citep{aime}, HMMT 2024, and HMMT 2025 ~\citep{hmmt_feb_2024_archive,hmmt_feb_2025_archive}. 
We additionally evaluate on GPQA-Diamond \citep{rein2023gpqagraduatelevelgoogleproofqa} to test generalization to graduate-level scientific reasoning resistant to pattern matching.
\paragraph{Baselines \& Variants.}
We compare against five inference-scaling baselines: unweighted \textbf{Self-Consistency (SC)} \citep{wang2023selfconsistencyimproveschainthought}, \textbf{Slim-SC} \citep{hong-etal-2025-slim} (which aggregates via cosine-similarity deduplication at a 0.95 threshold), \textbf{DeepConf} \citep{fu2025deepthinkconfidence} (which early-stops traces falling below an offline-calibrated top-10\% confidence threshold with 16 traces for initialization pool), and \textbf{STEP} \citep{liang2026hiddenstatesearlysignals} which prunes the lowest-scoring trace triggered by GPU memory constraint.
To isolate our algorithmic contributions, \sysname{} in our main evaluation uses the exact same off-the-shelf 2-layer MLP scorer from STEP. Moreover, we trained our own history-aware scorer in \Cref{sec:scorer} and demonstrate the generalization of \sysname{} in \Cref{sec:scorer_generalization}. 
\paragraph{Implementation Details.}
All methods are implemented atop vLLM \citep{kwon2023efficientmemorymanagementlarge} and evaluated on a single 275\,GB NVIDIA B300 GPU. 
Each problem is allocated a strict budget of $N = 256$ complete traces. 
For \sysname{}, tournament hyperparameters are held constant across all benchmarks and models: capacity $C{=}256$, swap size $K{=}16$, check interval $\Delta{=}200$ tokens, warmup threshold $w{=}12{,}000$ tokens, and hard-floor $\delta{=}0.1$. Final answers are aggregated via score-weighted majority vote, utilizing a position-weighted penalty that favors confidence in later generation steps. We have a detailed ablation study in \Cref{sec:hyper_ablation} to show the robustness of \sysname{} on hyperparameters.

\input{figures/experiment/main_results}
\subsection{Main Evaluation}
\Cref{tab:main_results} reports the end-to-end task accuracy and token consumption across all evaluated benchmarks. Because \sysname{} utilizes the identical baseline MLP scorer as STEP, any variance in accuracy or efficiency is strictly attributable to the underlying search topology.

\paragraph{Beam search strictly dominates baselines with substantial token efficiency.}
Replacing threshold-based pruning with our active beam search yields universal accuracy gains across all benchmarks and model architectures from 4B to 14B. For the 4B model, \sysname{} matches the highly calibrated DeepConf baseline on AIME-25 (90.0\%) while significantly outperforming it on HMMT-24 (+6.7\%) and GPQA (+2.6\%). When compared directly to the pruning-only STEP baseline, \sysname{} provides consistent lifts: +3.3\% on AIME-25 and HMMT-24 for the 4B model, +2.5\% on AIME-25 for the 8B model, and +1.6\% on HMMT-24 for the Phi-4 model. These improvements validate our core hypothesis: by continuously replenishing the trace pool via branching rather than allowing concurrency to passively decline, the tournament dynamically shifts the ensemble distribution toward correct regions of the search space, successfully breaking the majority-vote accuracy ceiling.

\input{figures/experiment/efficiency_accuracy_tradeoff}
\subsection{Efficiency: Token Consumption, Latency, and Trace Throughput}
\label{sec:efficiency}
We analyze system efficiency by quantifying absolute token consumption and hardware utilization over time with implementation of our system on vLLM ~\citep{kwon2023efficientmemorymanagementlarge}.
\input{figures/experiment/productive_throughput}
Combined with \Cref{tab:main_results}, \sysname{} establishes a new tradeoff frontier for efficiency (\Cref{fig:efficiency_tradeoff}). Across all evaluated architectures, our method achieves strict token reductions relative to traditional parallel sampling. 
These reductions are particularly pronounced on the Phi-4 architecture, where \sysname{} decreases the total token budget by 68.5\% on HMMT-25 (1.75M tokens versus SC's 5.56M). Similarly, on the Qwen3-4B model, we observe a 60.6\% token reduction on HMMT-24. This efficiency is a direct architectural consequence of thought-level branching: by forcing child traces to inherit the physical KV-cache of high-scoring parents, the marginal computational cost of exploring divergent reasoning pathways is vastly diminished.
In terms of latency, \sysname{} achieves more than \textbf{2$\times$ faster than parallel sampling} and \textbf{1.5$\times$ speedup other subtractive baselines}, while remaining competitive with STEP in raw latency. 
Although STEP can appear slightly faster (see \Cref{fig:efficiency_tradeoff}), it suffers from hardware underutilization: terminated traces are not replaced, causing the active batch to decay and shrinking the effective vote size. 
In contrast, \sysname{} continuously resupplies the trace pool via thought-level branching, maintaining full hardware utilization and yielding over \textbf{2$\times$ higher trace throughput} (e.g., 0.216 vs.\ 0.098 on Qwen3-4B; \Cref{fig:completion_rate}).

%% file: figures/experiment/main_results.tex
\vspace{-1em}
\begin{table*}[t]
\centering
\footnotesize
\newcommand{\sdt}[1]{{\scriptsize\,(#1)}}
\renewcommand{\arraystretch}{1.05}
\setlength{\tabcolsep}{3.5pt}
\caption{End-to-end task accuracy and token consumption per question ($\times 10^6$) at $N=256$. $\Delta$: relative token reduction (\%) vs.\ SC@256. \textbf{Bold}: best accuracy / lowest tokens per dataset.}
\label{tab:main_results}
\begin{tabular}{l cccccccccc}
\toprule
\multirow{2}{*}{\textbf{Method}} & \multicolumn{2}{c}{\textbf{AIME-25}} & \multicolumn{2}{c}{\textbf{AIME-26}} & \multicolumn{2}{c}{\textbf{HMMT-24}} & \multicolumn{2}{c}{\textbf{HMMT-25}} & \multicolumn{2}{c}{\textbf{GPQA}} \\
\cmidrule(lr){2-3} \cmidrule(lr){4-5} \cmidrule(lr){6-7} \cmidrule(lr){8-9} \cmidrule(lr){10-11}
& \textbf{Tok.}\sdt{$\Delta$} & \textbf{Acc} & \textbf{Tok.}\sdt{$\Delta$} & \textbf{Acc} & \textbf{Tok.}\sdt{$\Delta$} & \textbf{Acc} & \textbf{Tok.}\sdt{$\Delta$} & \textbf{Acc} & \textbf{Tok.}\sdt{$\Delta$} & \textbf{Acc} \\
\midrule
\multicolumn{11}{l}{\textit{\textbf{Qwen3-4B-Thinking}}} \\
SC@256         & 6.02 & 86.7 & 5.82 & 86.7 & 7.62 & 50.8 & 6.86 & 65.0 & 2.28 & 68.2 \\
Slim-SC        & 3.93\sdt{-34.7} & 86.7 & 3.83\sdt{-34.2} & 85.0 & 4.01\sdt{-47.4} & 51.7 & 3.72\sdt{-45.8} & 65.8 & 1.66\sdt{-27.2} & 65.6 \\
DeepConf       & 3.27\sdt{-45.7} & \textbf{90.0} & \textbf{3.19\sdt{-45.2}} & 86.7 & 4.28\sdt{-43.8} & 58.3 & 4.15\sdt{-39.5} & 66.7 & \textbf{1.52\sdt{-33.3}} & 67.6 \\
STEP           & 3.96\sdt{-34.2} & 86.7 & 3.51\sdt{-39.7} & 87.5 & 4.02\sdt{-47.2} & 61.7 & 3.88\sdt{-43.4} & 66.7 & 2.09\sdt{-8.3}  & 66.9 \\
\rowcolor{ourscolor}
\sysname{} (Ours) & \textbf{3.07\sdt{-49.0}} & \textbf{90.0} & 3.45\sdt{-40.7} & \textbf{88.3} & \textbf{3.00\sdt{-60.6}} & \textbf{65.0} & \textbf{3.49\sdt{-49.1}} & \textbf{67.5} & 1.83\sdt{-19.7} & \textbf{70.2} \\
\midrule
\multicolumn{11}{l}{\textit{\textbf{DeepSeek-R1-8B}}} \\
SC@256         & 6.76 & 83.3 & 6.85 & 83.3 & 8.47 & 55.8 & 7.65 & 70.0 & 2.92 & 67.1 \\
Slim-SC        & 5.87\sdt{-13.2} & 83.3 & 6.08\sdt{-11.2} & 83.3 & 7.32\sdt{-13.6} & 55.8 & 6.93\sdt{-9.4}  & 70.7 & 2.26\sdt{-22.6} & 67.1 \\
DeepConf       & 3.75\sdt{-44.5} & 81.7 & 3.67\sdt{-46.4} & 84.2 & 4.34\sdt{-48.8} & 60.0 & 3.97\sdt{-48.1} & 71.7 & \textbf{1.68\sdt{-42.5}} & \textbf{68.7} \\
STEP           & \textbf{3.71\sdt{-45.1}} & 83.3 & \textbf{3.66\sdt{-46.6}} & 83.3 & 4.15\sdt{-51.0} & 63.3 & 4.08\sdt{-46.7} & 75.8 & 2.44\sdt{-16.4} & 68.2 \\
\rowcolor{ourscolor}
\sysname{} (Ours) & 4.21\sdt{-37.7} & \textbf{85.8} & 4.09\sdt{-40.3} & \textbf{85.0} & \textbf{3.05\sdt{-64.0}} & \textbf{65.6} & \textbf{3.81\sdt{-50.2}} & \textbf{75.8} & 2.21\sdt{-25.3} & 68.2 \\
\midrule
\multicolumn{11}{l}{\textit{\textbf{Microsoft-Phi}}} \\
SC@256         & 4.24 & 86.7 & 4.20 & 88.5 & 5.43 & 56.7 & 5.56 & 73.3 & 3.05 & 76.3 \\
Slim-SC        & 3.43\sdt{-19.1} & 85.0 & 3.50\sdt{-16.7} & 86.7 & 4.72\sdt{-13.1} & 55.7 & 4.48\sdt{-19.4} & 74.2 & 2.24\sdt{-26.6} & 72.3 \\
DeepConf       & 2.56\sdt{-39.6} & 85.8 & 2.75\sdt{-34.5} & 86.7 & 2.86\sdt{-47.3} & 57.5 & 3.02\sdt{-45.7} & 74.2 & 1.61\sdt{-47.2} & 74.8 \\
STEP           & 2.39\sdt{-43.6} & 87.5 & 2.44\sdt{-41.9} & \textbf{90.0} & 2.75\sdt{-49.4} & 56.7 & 2.78\sdt{-50.0} & 75.0 & 2.10\sdt{-31.1} & 76.7 \\
\rowcolor{ourscolor}
\sysname{} (Ours) & \textbf{1.76\sdt{-58.5}} & \textbf{88.3} & \textbf{1.86\sdt{-55.7}} & \textbf{90.0} & \textbf{1.72\sdt{-68.3}} & \textbf{58.3} & \textbf{1.75\sdt{-68.5}} & \textbf{75.8} & \textbf{1.59\sdt{-47.9}} & \textbf{77.1} \\
\bottomrule
\end{tabular}
\end{table*}

%% file: figures/experiment/efficiency_accuracy_tradeoff.tex
\begin{figure}[!h]
    \centering
    \includegraphics[width=0.95\linewidth]{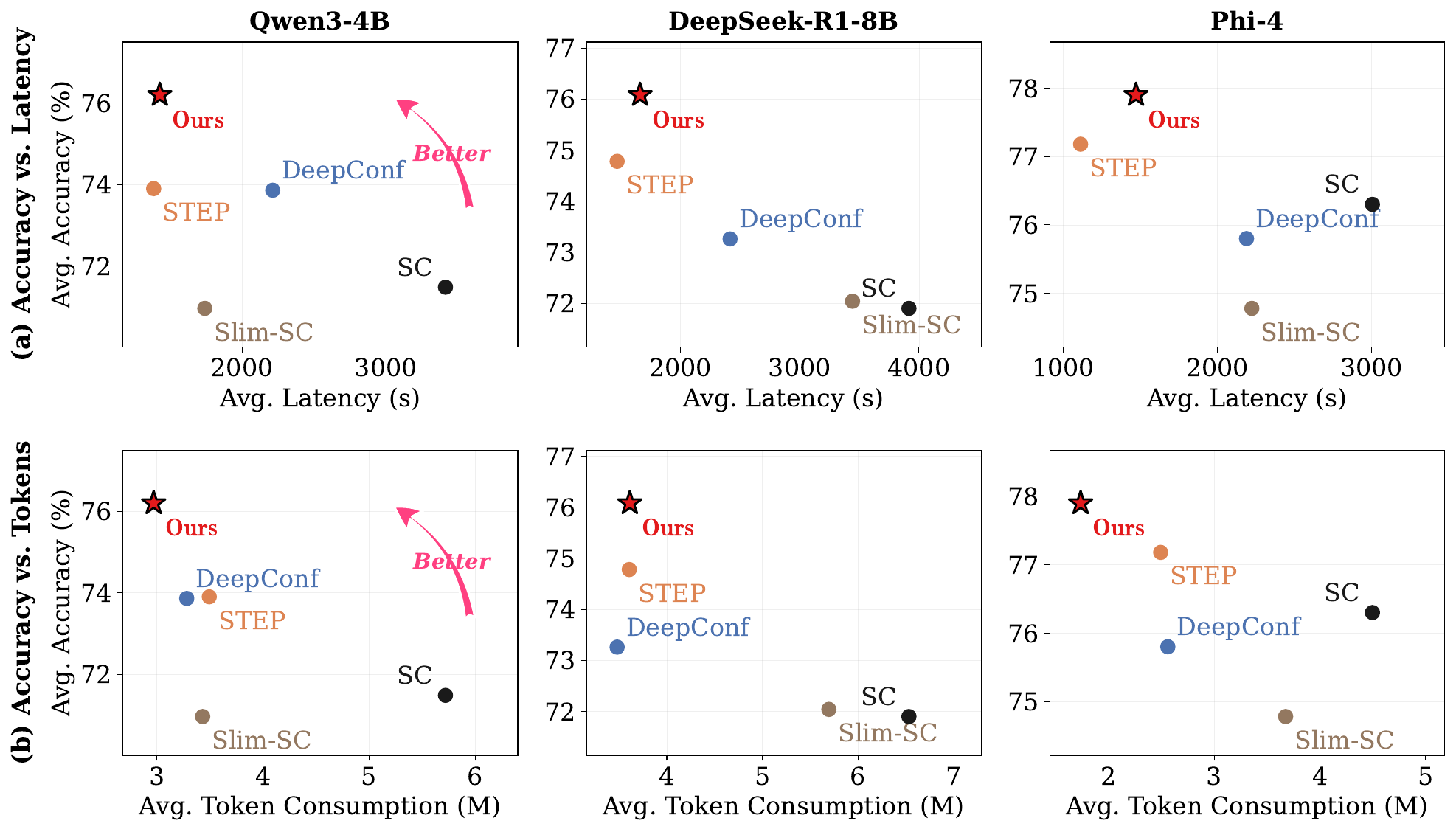}
    \caption{Efficiency vs. Accuracy trade-offs over all benchmarks. (a) Average Latency vs. Accuracy. (b) Average Token Consumption vs. Accuracy. \sysname{} (Ours) dominates the efficiency frontier, achieving the highest accuracy with the lowest overall token footprint and highly competitive wall-clock latency.}
    \label{fig:efficiency_tradeoff}
\end{figure}

%% file: figures/experiment/productive_throughput.tex

\begin{wrapfigure}{r}{0.4\columnwidth}
    \centering
    \includegraphics[width=\linewidth]{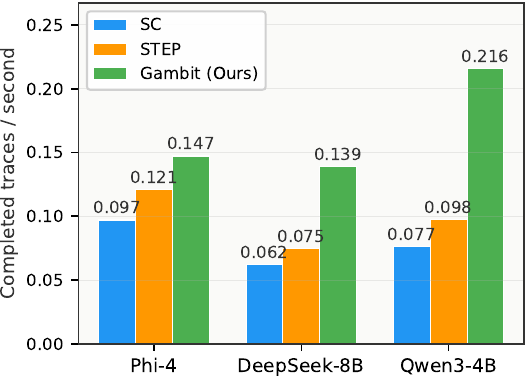}
    \caption{Trace throughput on AIME-26. \sysname{} effectively doubles the completion rate.}
    \label{fig:completion_rate}
\end{wrapfigure}

%% file: sections/conclusion.tex
\section{Analysis}
\label{sec:analysis}

\subsection{Generation Distribution and Decoding Dynamics}
\label{sec:generation_distribution}
\input{figures/experiment/analysis/length_distribution}

The empirical results in \Cref{sec:efficiency} reveal a non-trivial systems dynamic: \sysname{} achieves substantial reductions in total token consumption (up to 68.5\%), yet its wall-clock latency remains comparable to aggressive pruning methods such as STEP. To explain this apparent discrepancy, we analyze the distribution of generated tokens and trace lengths on AIME-2026.

\paragraph{Prefix Reuse and Unique Token Minimization.}
The reduction in total token consumption is driven by prefix reuse through branching. \Cref{fig:trace_len_unique} shows the distribution of \emph{unique} tokens generated per completed trace, defined as the total generated tokens excluding inherited prefix tokens. Because \sysname{} branches from shared high-quality prefixes in the KV-cache, it avoids recomputing early reasoning steps. For example, on Phi-4, \sysname{} generates a median of 5.2K unique tokens per trace, compared to 14.5K for parallel sampling (SC) and 7.7K for STEP. This demonstrates that \sysname{} explores alternative reasoning paths with substantially lower marginal cost, tightly controlling the total token budget.

\paragraph{Distribution Shift in Total Sequence Length.}
We next examine the \emph{total} sequence length of completed traces (\Cref{fig:trace_len_total}). At first glance, \sysname{} appears to produce significantly longer traces (e.g., median 35.8K vs.\ 16.5K for STEP on DeepSeek-R1-8B). However, this does \emph{not} indicate that individual reasoning trajectories are intrinsically longer under \sysname{}. Instead, this effect arises from a shift in the \emph{distribution} of surviving traces. Subtractive methods such as STEP terminate low-scoring traces without replacement, truncating a large fraction of trajectories early and shifting the length distribution toward shorter sequences. In contrast, \sysname{} both prunes and \emph{replenishes} the pool by branching from high-quality prefixes. 

Crucially, branching does not make individual traces longer; rather, it increases the \emph{frequency} of traces that continue to generate tokens. As a result, the overall distribution shifts to the right, even though the per-trace generation dynamics remain unchanged.

\paragraph{Implications for Decoding Latency.}
This distributional shift explains why token savings do not directly translate into proportional latency reductions. Although \sysname{} minimizes redundant token generation via prefix sharing, it allocates more compute to continuing high-quality trajectories, resulting in a higher fraction of long-running traces.

Overall, \sysname{} achieves a favorable trade-off: it dramatically reduces total token consumption while maintaining competitive wall-clock latency by reallocating compute toward promising trajectories rather than uniformly truncating them.

\section{Conclusion}
\label{sec:conclusion}
\sysname{} is a thought-level beam search framework that formulates test-time reasoning as a constrained compute-allocation problem over partial trajectories. Rather than scaling inference by increasing the number of independent samples, our approach dynamically reallocates compute toward promising intermediate states while maintaining strict hardware constraints. Our results demonstrate that this shift is both algorithmically and systemically necessary. By actively branching from high-quality prefixes and enforcing a zero-sum allocation policy, it simultaneously improves accuracy and efficiency—achieving consistent gains across benchmarks while reducing token consumption by up to 68.5\% and maintaining high hardware utilization. Empirically, this leads to strict dominance over both parallel sampling and pruning-based approaches.

%% file: figures/experiment/analysis/length_distribution.tex
\begin{figure*}[h]
    \centering
    \includegraphics[width=\textwidth]{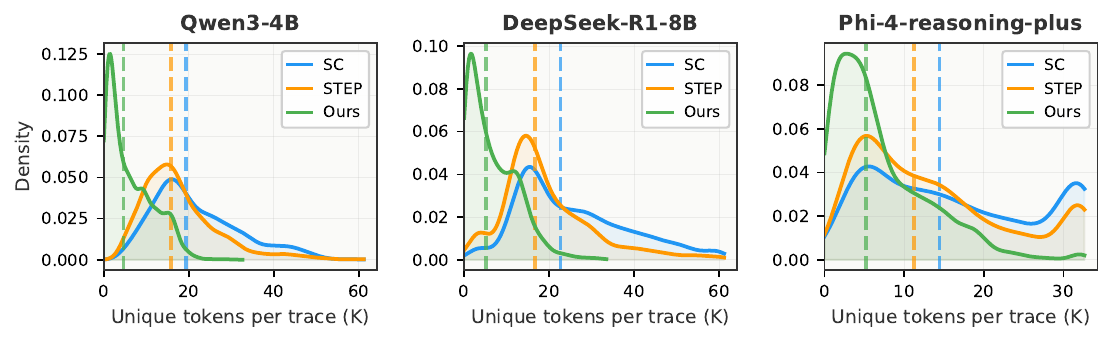}
    \caption{Distribution of \emph{unique} tokens generated per completed trace on AIME-2026. By branching from shared high-quality KV-cache prefixes, \sysname{} drastically reduces the median number of new tokens required to explore alternative reasoning pathways compared to SC and STEP.}
    \label{fig:trace_len_unique}
    
    \vspace{1.5em} 
    
    \includegraphics[width=\textwidth]{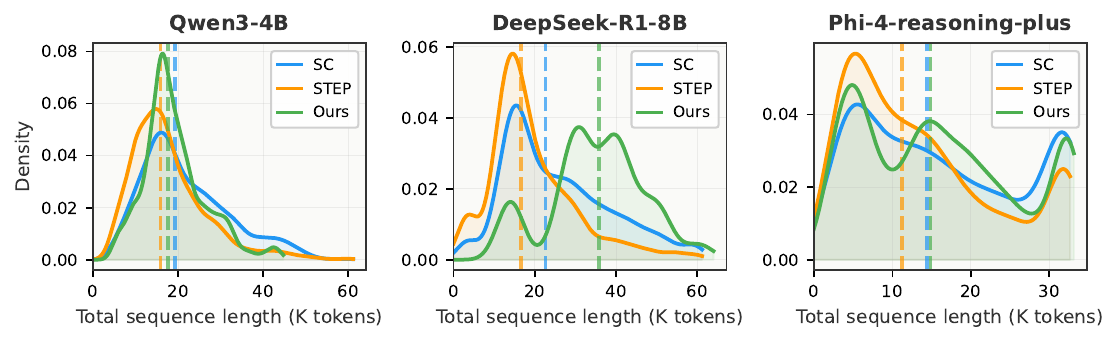}
    \caption{Distribution of \emph{total} sequence length (including inherited prefixes) for completed traces on AIME-2026. \sysname{} sustains significantly deeper, rigorous reasoning chains than subtractive baselines like STEP, which accounts for the serial decoding latency bottleneck despite massive overall token savings.}
    \label{fig:trace_len_total}
\end{figure*}

%% file: sections/appendix.tex
\subsection{Evaluating Search Generalization: A History-Aware Sequence Scorer }
To evaluate how different scoring signals affect search dynamics, we have trained our history-aware sequence model (\Cref{sec:scorer}).
\label{sec:scorer}
The proposed search algorithm is agnostic to the specific scoring function used to evaluate trajectories. However, to study how the quality of the scoring signal affects search behavior (\Cref{sec:scorer_generalization}), we design an alternative, history-aware \emph{sequence scorer} $f_\theta$ to off-the-shelf scorers used in main evaluation. 
Unlike a Markovian MLP, this sequence scorer maps a prefix of hidden-state vectors to a scalar quality score in $[0,1]$ by attending to the full reasoning trajectory, providing a structurally distinct guidance signal for our ablations.

\paragraph{Data Collection \& Step-Level States.}
We generate $n$ independent traces per problem using standard parallel sampling. Each trace is delimited by double-newline boundaries (\texttt{\textbackslash n\textbackslash n}) within the model's \texttt{<think>} block~\citep{pan2025specreason}, with a binary correctness label $y \in \{0, 1\}$ assigned to the final extracted answer. During a single forward pass, we extract the last-layer hidden state $\mathbf{h}_t \in \mathbb{R}^{D}$ at the token immediately preceding each boundary~\citep{liang2026hiddenstatesearlysignals}. This yields a per-trace step sequence $(\mathbf{h}_1, \mathbf{h}_2, \ldots, \mathbf{h}_T)$.

\paragraph{Scorer Architecture.}
The sequence scorer $f_\theta$ is a compact Transformer:
\begin{align}
    \tilde{\mathbf{h}}_t &= \mathrm{GELU}\!\left(\mathbf{W}_{\mathrm{in}} \cdot \mathrm{LN}(\mathbf{h}_t)\right), \label{eq:proj} \\
    \mathbf{z}_1, \ldots, \mathbf{z}_T &= \mathrm{SequenceTransformer}(\tilde{\mathbf{h}}_1, \ldots, \tilde{\mathbf{h}}_T), \label{eq:transformer} \\
    \hat{y}_t &= \sigma\!\left(\mathbf{w}_{\mathrm{out}}^\top \cdot \mathrm{LN}(\mathbf{z}_t)\right). \label{eq:head}
\end{align}
Each layer utilizes multi-head causal self-attention with Rotary Position Embeddings (RoPE)~\citep{su2023roformerenhancedtransformerrotary} and SwiGLU feed-forward networks~\citep{shazeer2020gluvariantsimprovetransformer}. The causal mask is essential: position $t$ attends to all prior steps $1, \ldots, t$, ensuring $\hat{y}_t$ is conditioned on the full reasoning history up to step $t$. This allows the sequence scorer to detect globally inferior steps that appear locally plausible—providing a high-resolution signal to test \sysname{}'s amplification dynamics.

\paragraph{Training Objective.}
Because the sequence scorer is queried dynamically during inference via the logit at the final observed step, we train exclusively with a \textbf{last-step} binary cross-entropy loss:
\begin{equation}
    \mathcal{L}(\theta) = -\frac{1}{|\mathcal{D}|} \sum_{(\mathbf{h}_{1:T},\, y) \in \mathcal{D}} \left[ y \log \hat{y}_T + (1-y) \log (1 - \hat{y}_T) \right], \label{eq:bce}
\end{equation}
where $T$ is the sequence length and $\hat{y}_T$ is the output at the last valid position. Crucially, only the final position contributes a gradient signal, forcing the attention mechanism to actively credit-assign the full trajectory context rather than relying on local features at any single step.

\subsection{Generalization Across Scoring Architectures}
\input{figures/experiment/ranking_accuracy}
\label{sec:scorer_generalization}
\input{figures/experiment/seqscorer_eval}
A fundamental theoretical difference between subtractive pruning (STEP) and active compute reallocation (\sysname{}) is how they utilize the underlying guidance signal. In a pruning-only system, the scoring mechanism acts as a negative filter: it can terminate unpromising traces, but it cannot actively construct or discover correct ones. In contrast, \sysname{} treats the scoring mechanism as a generative compass, using it to systematically explore promising regions of the solution space. By actively branching from high-scoring prefixes, \sysname{} structurally shifts the sampling distribution toward correct answers.

To verify that \sysname{}'s algorithmic superiority generalizes beyond the single, off-the-shelf MLP scorer used in our main evaluation, we evaluate both inference algorithms using our custom, history-aware sequence scorer (\Cref{sec:scorer}). This provides a distinct guidance signal derived from the full latent trajectory rather than isolated steps.

As shown in \Cref{tab:scorer_ablation}, replacing the baseline MLP with the trajectory-aware scorer reveals a stark contrast between the two search topologies:

\paragraph{Active search robustly dominates subtractive pruning.}
Across all base models and benchmarks, \sysname{} robustly and significantly outperforms the STEP baseline. On the 8B model, \sysname{} achieves substantial end-to-end accuracy gains over STEP, including +7.7\% on HMMT-24 and +5.2\% on AIME-25. Similarly, on the 4B model, \sysname{} maintains consistent superiority, delivering a +3.3\% lift on AIME-25 and +1.7\% on HMMT-24. This confirms that our tournament algorithm's advantages are fundamentally robust and architecture-agnostic; its compute reallocation mechanism consistently yields higher downstream task accuracy regardless of the underlying reward model.

\paragraph{Subtractive systems bottleneck guidance signals.}
The magnitude of these performance gaps (e.g., +7.7\% on HMMT-24) highlights a critical limitation of pruning-only frameworks. Even when provided with alternative guidance signals, STEP is fundamentally bottlenecked by its inability to actively replenish the trace pool. If the base model fails to generate a sufficient density of correct traces organically, a filter merely results in a smaller, equally biased pool of survivors. 
Conversely, \sysname{} fully operationalizes the provided quality signal. By actively branching from prefixes identified as highly promising, the algorithm physically reshapes the output distribution, translating the guidance signal into robust, state-of-the-art benchmark wins without being hypersensitive to the specific choice of scorer.

\subsection{Latency Decomposition and System Overhead}
\label{sec:cost_profiling}

A common vulnerability in complex inference-time search algorithms is the introduction of prohibitive CPU-bound bookkeeping or memory-management bottlenecks. To validate that our thought-level beam search imposes minimal structural overhead, we decompose the end-to-end wall-clock latency of \sysname{}.

\begin{table}[h]
\centering
\small
\renewcommand{\arraystretch}{1.15}
\caption{Latency decomposition for \sysname{} with Qwen3-4B on AIME-26  with a batch size of $N=256$. Algorithmic overhead accounts for less than 1\% of total execution time, confirming that the beam search mechanism operates efficiently without introducing system bottlenecks.}
\label{tab:latency_decomposition}
\begin{tabular}{lrr}
\toprule
\textbf{Execution Component} & \textbf{Time (s)} & \textbf{\% Wall Clock} \\
\midrule
GPU Compute (Forward Pass \& Sampling)     & 37,851.97 & 99.03\% \\
Beam-search Overhead (Scoring \& Tree Mgmt) & 370.76    & 0.97\% \\
\midrule
\textbf{Total Wall Clock}                  & \textbf{38,222.73} & \textbf{100.00\%} \\
\bottomrule
\end{tabular}
\end{table}

As detailed in \Cref{tab:latency_decomposition}, the operational overhead introduced by the tournament mechanism is strictly negligible. The core autoregressive forward passes and GPU sampling routines completely dominate execution, consuming 99.03\% of the total wall-clock time. 

The remaining fraction of execution time (0.97\%) encompasses all algorithmic interventions required to dynamically manage the search space. The majority of this overhead ($\sim$0.60\% of total time) is attributed to \emph{guidance synchronization}—the inter-process communication (RPC) required to transmit latent hidden states to the reward model worker and retrieve trajectory scores. This functions primarily as a brief synchronization stall rather than a computational bottleneck. \emph{Tree topology management} accounts for approximately 0.25\%, encapsulating the CPU-bound bookkeeping necessary to sort active traces and dispatch prune-and-branch commands. Finally, \emph{continuous scheduling and garbage collection} ($\sim$0.12\%) manages the per-step priority queues and hard-floor threshold scans, ensuring that degenerated traces are opportunistically evicted without interrupting the continuous batching cycle.

Ultimately, this profiling confirms that highly dynamic, thought-level beam search topologies can be executed with virtually no systems penalty. By effectively mitigating CPU and memory-management overheads, \sysname{} ensures that the efficiency gains achieved through token reductions translate directly into real-world latency improvements.

\subsection{Hyperparameter Ablations}
\label{sec:hyper_ablation}
\input{figures/experiment/analysis/hyperparameter_ablation}

We analyze the sensitivity of \sysname{} to key design parameters, including swap size $K \in \{4, 8, 16, 32\}$, beam search interval $\Delta \in \{100, 200, 400\}$, warmup threshold $w \in \{8\mathrm{K}, 12\mathrm{K}, 16\mathrm{K}\}$, and GPU memory usage ratio $\delta$ (Figure~\ref{fig:ablation}). 
Across all dimensions, performance varies smoothly without sharp degradation, indicating that \sysname{} is robust to hyperparameter choices. 
The selected configuration $(K{=}16, \Delta{=}200, w{=}12\mathrm{K}, r{=}0.9)$ lies in a broad high-performing region and consistently outperforms the SC@256 baseline.

Each parameter reflects a balance between accuracy and efficiency. 
Small swap sizes limit the ability to refresh low-quality traces, while overly large values can disrupt stable decoding; $K{=}16$ provides a consistent improvement. 
Frequent updates ($\Delta{=}100$) act on insufficiently developed prefixes, while infrequent updates ($\Delta{=}400$) delay corrective selection; $\Delta{=}200$ yields a stable cadence. 
Similarly, branching too early can harm reasoning quality, whereas excessive delay wastes compute, and a moderate warmup ($w{=}12\mathrm{K}$) allows traces to establish a reliable foundation before selection. 
Finally, increasing the GPU memory usage ratio $r$ improves performance, highlighting the importance of effective KV-cache utilization for parallel decoding. 
Overall, \sysname{} operates in a stable regime where moderate hyperparameters achieve strong performance while maintaining efficient and well-conditioned decoding dynamics.

\subsection{Runtime Example of \sysname{}}
\label{app:runtime-example}

To make the runtime behavior of \sysname{} concrete, we open up a single
tournament tree and walk through four representative traces it produced
on AIME 2025 Q12.  The configuration is the standard one used
throughout Section~5: capacity $C{=}256$, swap size $K{=}16$, check
interval $\Delta{=}200$, warmup $w{=}12{\rm K}$, hard floor
$\delta{=}0.1$, off-the-shelf MLP scorer $f_\theta$ from STEP, and
model DeepSeek-R1-0528-Qwen3-8B.  The problem asks for the expected
number of regions when two perpendicular diameters and $25$ random
chords (with endpoints in different quadrants) are drawn through a
disk; the ground-truth answer is $204$.

\paragraph{Tree at a glance.}
The run produced a tournament tree with $1{,}595$ nodes:
\begin{itemize}
\setlength{\itemsep}{0pt}
\item $256$ root traces seeded at the start of decoding;
\item $1{,}339$ branches spawned over the run by zero-sum prune-and-branch
      operations (Algorithm~1);
\item $1{,}083$ traces eventually \emph{pruned}; their partial outputs are
      not preserved in the tree export;
\item $256$ traces eventually \emph{completed}, contributing answers to
      the score-weighted majority vote;
\item $42$ of the $256$ completed traces returned the correct answer
      $204$; the score-weighted vote selects $204$ with
      $\sum \bar s_\tau = 24.02$ versus $21.81$ for the runner-up.
\end{itemize}

\paragraph{Baseline contrast: parallel sampling at twice the trace
budget cannot solve this question.}
For comparison, on the same problem, with the same model, plain
self-consistency at $n{=}512$ (no scoring, no pruning, no branching;
$\delta\!=\!\text{none}$, every trace runs to completion in
parallel) consumes $21$ million tokens over $15K$ seconds of
wall-clock time --- roughly $13\times$ the new tokens and
$3.4\times$ the wall clock of the \sysname{} run above --- and
nonetheless fails to recover the correct answer.  Of its $512$
completed traces, only $34$ ($6.6\%$) return $204$, while the
incorrect answer $\tfrac{487}{3}$ wins the unweighted plurality with
$90$ votes ($17.6\%$); $\tfrac{637}{3}$ comes second with $59$
($11.5\%$) and $204$ comes only third. With the same setup,
allocated by \sysname{} via thought-level beam search, \sysname{} both
delivers the correct answer and uses an order of magnitude fewer
generated tokens.

\paragraph{The subtree we follow.}
We pick one root prefix $\tau_0$ that survived past the warmup window.
Its surviving depth-1 descendant, denoted $\tau_p$, became the
\emph{highest-scoring} depth-1 trace in the entire run with running
score $\bar s_{\tau_p}{=}0.610$.  Each tournament round in which
$\tau_p$ ranks in the top-$K$ executes a single
\emph{branching event}: $\tau_p$'s prefix is shared into two
trajectories, one of which is $\tau_p$ continuing decoding, and the
other is a fresh depth-2 child that begins its own continuation from
the same prefix via prefix-cache reuse.  We follow three such
branching events for $\tau_p$, each producing one depth-2 child of a
different eventual fate.  Schematically:
\begin{quote}\small\ttfamily\sloppy
$\tau_0$\ (root, depth 0)\\
\hspace*{1.6em}\textbar\\
\hspace*{1.6em}$+$ $\tau_p$\ (depth 1; completed at decoding step\ $\sim$34{,}000;\\
\hspace*{1.6em}\hspace*{1.6em}\hspace*{1.6em}$\bar s=0.610$, ans = 203)\\
\hspace*{1.6em}\hspace*{1.6em}\textbar\\
\hspace*{1.6em}\hspace*{1.6em}\textbar---\ branching event at decoding step\ $\sim$26{,}500\ \\
\hspace*{1.6em}\hspace*{1.6em}\textbar\hspace*{2.4em}$+$ $\tau_C$\ (depth 2; completed; $\bar s=0.566$,\\
\hspace*{1.6em}\hspace*{1.6em}\textbar\hspace*{2.4em}\hspace*{1.6em}\hspace*{1.6em}ans = 559/3)\ \ ---\ alternate-explore child\\
\hspace*{1.6em}\hspace*{1.6em}\textbar\\
\hspace*{1.6em}\hspace*{1.6em}\textbar---\ branching event at decoding step\ $\sim$26{,}000\ \\
\hspace*{1.6em}\hspace*{1.6em}\textbar\hspace*{2.4em}$+$ $\tau_D$\ (depth 2; pruned in a later round;\\
\hspace*{1.6em}\hspace*{1.6em}\textbar\hspace*{2.4em}\hspace*{1.6em}\hspace*{1.6em}no preserved text)\ \ ---\ pruned child\\
\hspace*{1.6em}\hspace*{1.6em}\textbar\\
\hspace*{1.6em}\hspace*{1.6em}\textbar---\ branching event at decoding step\ $\sim$30{,}000\ \\
\hspace*{1.6em}\hspace*{1.6em}\hspace*{2.4em}$+$ $\tau_B$\ (depth 2; completed; $\bar s=0.632$,\\
\hspace*{1.6em}\hspace*{1.6em}\hspace*{2.4em}\hspace*{1.6em}\hspace*{1.6em}ans = 204)\ \ ---\ correct child
\end{quote}
\noindent
Each branching event is a single 1-to-1 split of $\tau_p$'s prefix:
$\tau_p$ continues, one new depth-2 child begins.  The three events
above are the ones whose products we reproduce below; $\tau_p$ takes
part in additional branching events at later tournament rounds, all
following the same 1-to-1 pattern.  We now reproduce the four marked
traces $\tau_p$, $\tau_B$ verbatim, each with
metadata followed by the model's relevant output.
``\texttt{[\dots]}'' marks elisions for length.

\subsubsection*{Trace $\tau_p$ --- the high-scoring depth-1 founder
(completed, answer $203$, $\bar s = 0.610$)}

\noindent\emph{Metadata.}
Branching depth $1$; created at decoding step $13{,}934$ as a branch
from root $\tau_0$; generated $20{,}669$ new tokens after inheriting
$14{,}004$ tokens from $\tau_0$ via prefix-cache reuse; participated
in several subsequent branching events as the high-scoring parent
$\tau_p$ (three of which we follow below); final completion at
decoding step $\sim 34{,}000$ with extracted answer $203$.

\noindent\emph{Excerpt --- opening.}

\begin{quote}\small\sloppy
\begin{verbatim}
<think>
Alex divides a disk into four quadrants with two perpendicular
diameters.  So, that's like the x and y axes, dividing the disk
into four equal parts.  Then he draws 25 more lines by connecting
two random points on the perimeter in different quadrants.  ...
And we need to find the expected number of regions that these 27
lines divide the disk into.
\end{verbatim}
\end{quote}

\noindent\texttt{[\dots\ the trace sets up linearity of expectation,
derives the chord-length distribution and the chord-vs-quadrant
configuration, integrates to obtain the chord-chord intersection
probability, and arrives at $E[N_i] = 8$ new regions per chord. \dots]}

\noindent\emph{Excerpt --- final calculation (the slip).}

\begin{quote}\small\sloppy
\begin{verbatim}
For each of the 25 random chords, 8.

So total E[R] = 1 + 2 + 25 * 8 = 3 + 200 = 203.

[...]

Now, expected regions 203.

Now, I think that's the answer.
</think>
The total expected number of regions is the sum of the new regions
created by all lines: 3 from the diameters and 25 x 8 from the
random chords, giving 3 + 200 = 203.

\boxed{203}
\end{verbatim}
\end{quote}

\noindent
Note the slip in the very first line of the excerpt: $\tau_p$ writes
``$1 + 2 + 25 \cdot 8$'' but evaluates to $3 + 200 = 203$, treating
the leading $1$ and $2$ as the diameter contributions and forgetting
that the disk itself contributes one initial region.  The final
boxed answer is therefore off by one.

\subsubsection*{Trace $\tau_B$ --- depth-2 child of $\tau_p$
(completed, answer $\mathbf{204}$, $\bar s = 0.632$)}

\noindent\emph{Metadata.}
Branching depth $2$; spawned from $\tau_p$ at decoding step
$\sim\!30{,}000$ during a tournament round in which $\tau_p$ ranked
first; inherited $31{,}404$ tokens from $\tau_p$ via prefix-cache
reuse; generated $3{,}649$ new tokens of its own; running score
$\bar s_{\tau_B} = 0.632$, the highest among all $256$ completed
traces; extracted answer $204$.  Because the score is the largest,
$\tau_B$ contributes the single largest mass to the score-weighted
vote.

\noindent\emph{Last $\approx\!100$ characters of inherited prefix
(shared with $\tau_p$).}

\begin{quote}\small\sloppy
\begin{verbatim}
... in the diameters, we computed their contribution separately.
\end{verbatim}
\end{quote}

\noindent\emph{Excerpt --- start of $\tau_B$'s newly generated
continuation.}

\begin{quote}\small\sloppy
\begin{verbatim}
.

Now, another way to think: total expected regions is
E[R] = E[1 + sum_{all lines} new regions created]

But since the lines include the diameters and the random chords,
and we have computed for each.

[...]

Now, so total E[R] = E[new regions by diameters]
                   + sum_{i=1}^{25} E[new regions by C_i]
                   = 3 + 25 * 8 = 203.

Now, but is there any overlap or something? No, because each
new region is created by exactly one line.

Now, so it should be correct.

Now, but let me confirm with a smaller case.
\end{verbatim}
\end{quote}

\noindent\emph{Excerpt --- $\tau_B$ catches the parent's slip.}

\begin{quote}\small\sloppy
\begin{verbatim}
Suppose we have only the two diameters and no random chords.
Then E[R] = 4, but according to formula, diameters contribute 3,
no random chords, so 3, but should be 4, inconsistency.

Mistake here.

When we add the diameters, they create 3 new regions, but
initially there was 1 region, so total regions after diameters
is 1 + 3 = 4.

[...]

Ah, I see, the issue is that the initial region is not accounted
for in the sum.

In the standard formula, the total number of regions is equal to
1 (the initial whole space) plus the sum over all lines of the
number of new regions created by that line.

[...]

In our calculation for the random chords, we are only summing the
new regions created by each line, but not including the initial
region.

In the expression for E[R], it should be
       E[R] = E[1 + sum_{all lines} new regions created]
\end{verbatim}
\end{quote}

\noindent\emph{Excerpt --- final answer.}

\begin{quote}\small\sloppy
\begin{verbatim}
</think>
The disk is initially divided into four quadrants by two
perpendicular diameters, which create 4 regions.  ...

The two diameters create a total of 3 new regions ...

Each random chord, when added, creates a random number of new
regions, but the expected number of new regions created by each
random chord is 8.  ...

With 25 random chords, the sum of the expected new regions created
by all random chords is 25 x 8 = 200.

The sum of the expected new regions created by all lines is
3 + 200 = 203.

The total expected number of regions is the initial region plus
this sum: 1 + 203 = 204.

\boxed{204}
\end{verbatim}
\end{quote}

\noindent
$\tau_B$'s contribution is small in token count (only $3{,}649$ new
tokens, almost free thanks to prefix-cache reuse) but decisive in
content: the child explicitly performs a smaller-case sanity check,
identifies the missing $+1$ from the disk, and writes the corrected
equation $1 + 203 = 204$.  This is the prototypical case of
\emph{branching repairs the parent}.  No subtractive method
(STEP, DeepConf, Slim-SC) can produce this kind of repair --- they can
only kill the parent for being below threshold or produce confidently wrong answers.

\subsubsection*{Conclusion}

The four traces above span the full state space a trace can occupy
in a \sysname{} run.  $\tau_p$ is a long-lived high-scoring trace that
itself completes \emph{and} serves as a branching parent.  $\tau_B$
is a short, late-branching child whose only job is to refine its
parent's prefix; in this case the refinement is critical because
$\tau_p$ slipped on the closing addition. $\tau_B$ alone
contributes the largest single mass to that vote.

\newpage
\subsection{End-to-end Latency and Token Consumption}
The following \Cref{tab:efficiency} records the data of wall-clock time and total token consumption for each model and benchmark. 
\begin{table}[t]
\centering
\small
\caption{Systems-level efficiency at $n{=}256$ on a single B300-275\,GB.
Token consumption is reported in thousands (K); latency in wall-clock seconds per question.
\textbf{Bold} indicates the best result per column within each model--metric block.}
\label{tab:efficiency}
\setlength{\tabcolsep}{4.5pt}
\renewcommand{\arraystretch}{1.15}
\begin{tabular}{ll rrrrr}
\toprule
& & \textbf{AIME-25} & \textbf{AIME-26} & \textbf{HMMT-24} & \textbf{HMMT-25} & \textbf{GPQA} \\
\midrule
\multicolumn{7}{l}{\textit{Qwen3-4B-Thinking-2507}} \\[2pt]
\multirow{5}{*}{\rotatebox[origin=c]{90}{\scriptsize Tokens\,(K)}}
  & SC         & 6{,}023 & 5{,}817 & 7{,}622 & 6{,}857 & 2{,}276 \\
  & Slim-SC    & 3{,}930 & 3{,}829 & 4{,}010 & 3{,}723 & 1{,}659 \\
  & DeepConf   & 3{,}266 & \textbf{3{,}192} & 4{,}278 & 4{,}148 & \textbf{1{,}516} \\
  & STEP       & 3{,}960 & 3{,}506 & 4{,}023 & 3{,}884 & 2{,}089 \\
  & \cellcolor{ourscolor}\sysname{} (Ours) & \cellcolor{ourscolor}\textbf{3{,}071} & \cellcolor{ourscolor}3{,}449 & \cellcolor{ourscolor}\textbf{3{,}002} & \cellcolor{ourscolor}\textbf{3{,}493} & \cellcolor{ourscolor}1{,}826 \\
\cmidrule{2-7}
\multirow{5}{*}{\rotatebox[origin=c]{90}{\scriptsize Latency\,(s)}}
  & SC         & 3{,}465 & 3{,}338 & 5{,}358 & 4{,}313 &   593 \\
  & Slim-SC    & 1{,}905 & 1{,}841 & 2{,}412 & 1{,}995 &   555 \\
  & DeepConf   & 2{,}095 & 2{,}188 & 3{,}231 & 2{,}948 &   605 \\
  & STEP       & 1{,}390 & 1{,}232 & 2{,}016 & \textbf{1{,}507} &   786 \\
  & \cellcolor{ourscolor}\sysname{} (Ours) & \cellcolor{ourscolor}\textbf{1{,}350} & \cellcolor{ourscolor}\textbf{1{,}177} & \cellcolor{ourscolor}\textbf{1{,}912} & \cellcolor{ourscolor}2{,}179 & \cellcolor{ourscolor}\textbf{527} \\
\midrule
\multicolumn{7}{l}{\textit{DeepSeek-R1-0528-Qwen3-8B}} \\[2pt]
\multirow{5}{*}{\rotatebox[origin=c]{90}{\scriptsize Tokens\,(K)}}
  & SC         & 6{,}764 & 6{,}849 & 8{,}465 & 7{,}652 & 2{,}919 \\
  & Slim-SC    & 5{,}874 & 6{,}080 & 7{,}323 & 6{,}933 & 2{,}256 \\
  & DeepConf   & 3{,}753 & 3{,}666 & 4{,}337 & 3{,}972 & \textbf{1{,}679} \\
  & STEP       & \textbf{3{,}711} & \textbf{3{,}661} & 4{,}147 & 4{,}077 & 2{,}444 \\
  & \cellcolor{ourscolor}\sysname{} (Ours) & \cellcolor{ourscolor}4{,}211 & \cellcolor{ourscolor}4{,}089 & \cellcolor{ourscolor}\textbf{3{,}053} & \cellcolor{ourscolor}\textbf{3{,}806} & \cellcolor{ourscolor}2{,}211 \\
\cmidrule{2-7}
\multirow{5}{*}{\rotatebox[origin=c]{90}{\scriptsize Latency\,(s)}}
  & SC         & 4{,}078 & 4{,}107 & 5{,}640 & 4{,}873 &   880 \\
  & Slim-SC    & 3{,}462 & 3{,}564 & 5{,}071 & 4{,}342 &   771 \\
  & DeepConf   & 2{,}692 & 2{,}582 & 3{,}256 & 2{,}800 &   744 \\
  & STEP       & \textbf{1{,}519} & \textbf{1{,}469} & \textbf{1{,}902} & \textbf{1{,}715} &   730 \\
  & \cellcolor{ourscolor}\sysname{} (Ours) & \cellcolor{ourscolor}1{,}571 & \cellcolor{ourscolor}1{,}842 & \cellcolor{ourscolor}2{,}014 & \cellcolor{ourscolor}2{,}169 & \cellcolor{ourscolor}\textbf{702} \\
\midrule
\multicolumn{7}{l}{\textit{Phi-4}} \\[2pt]
\multirow{5}{*}{\rotatebox[origin=c]{90}{\scriptsize Tokens\,(K)}}
  & SC         & 4{,}243 & 4{,}198 & 5{,}426 & 5{,}559 & 3{,}050 \\
  & Slim-SC    & 3{,}426 & 3{,}503 & 4{,}716 & 4{,}482 & 2{,}242 \\
  & DeepConf   & 2{,}557 & 2{,}749 & 2{,}860 & 3{,}022 & 1{,}608 \\
  & STEP       & 2{,}390 & 2{,}435 & 2{,}754 & 2{,}780 & 2{,}097 \\
  & \cellcolor{ourscolor}\sysname{} (Ours) & \cellcolor{ourscolor}\textbf{1{,}756} & \cellcolor{ourscolor}\textbf{1{,}855} & \cellcolor{ourscolor}\textbf{1{,}717} & \cellcolor{ourscolor}\textbf{1{,}752} & \cellcolor{ourscolor}\textbf{1{,}590} \\
\cmidrule{2-7}
\multirow{5}{*}{\rotatebox[origin=c]{90}{\scriptsize Latency\,(s)}}
  & SC         & 2{,}760 & 2{,}636 & 3{,}848 & 4{,}046 & 1{,}730 \\
  & Slim-SC    & 2{,}123 & 2{,}166 & 3{,}069 & 2{,}704 & 1{,}048 \\
  & DeepConf   & 1{,}864 & 1{,}749 & 2{,}437 & 2{,}832 & 2{,}056 \\
  & STEP       & \textbf{1{,}155} & \textbf{1{,}139} & \textbf{1{,}197} & \textbf{1{,}222} & \textbf{846} \\
  & \cellcolor{ourscolor}\sysname{} (Ours) & \cellcolor{ourscolor}1{,}480 & \cellcolor{ourscolor}1{,}538 & \cellcolor{ourscolor}1{,}623 & \cellcolor{ourscolor}1{,}589 & \cellcolor{ourscolor}1{,}123 \\
\bottomrule
\end{tabular}
\end{table}

%% file: figures/experiment/ranking_accuracy.tex
\begin{figure}[h]               
    \centering   
    \includegraphics[width=\linewidth]{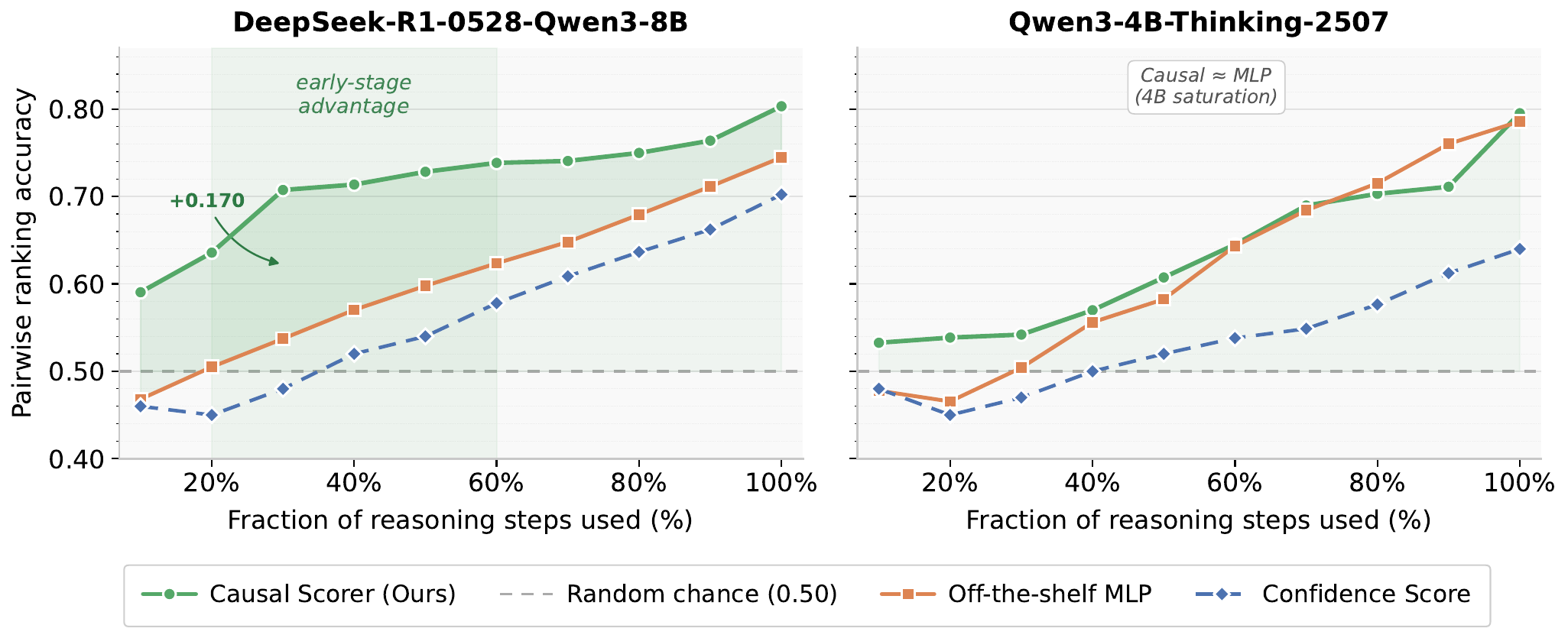}          
    \caption{\textbf{Pairwise ranking accuracy vs.\ fraction of reasoning steps used ($k\%$).}
    Higher accuracy indicates better separation of correct and incorrect traces given only a partial prefix.}                    
    \label{fig:ranking_accuracy}                            
\end{figure}

%% file: figures/experiment/seqscorer_eval.tex
\begin{table}[h]
\centering
\caption{Impact of scorer architecture on end-to-end downstream accuracy. \sysname{} consistently outperforms the pruning-only baseline (STEP) across different reward models and base LLMs. By actively reallocating compute to promising prefixes, \sysname{} effectively unlocks the potential of the guidance signal, whereas purely subtractive methods remain fundamentally bottlenecked by their inability to generate new traces.}
\label{tab:scorer_ablation}
\renewcommand{\arraystretch}{1.25}
\resizebox{\textwidth}{!}{%
\begin{tabular}{ll ccccc}
\toprule
\textbf{Base Model} & \textbf{System} & \textbf{AIME-25} & \textbf{AIME-26} & \textbf{HMMT-24} & \textbf{HMMT-25} & \textbf{GPQA} \\
\midrule
\multirow{2}{*}{\textit{\shortstack[l]{DeepSeek-R1-0528\\-Qwen3-8B}}}
& STEP + \scorername
& 83.3 & 84.2 & 61.7 & 75.8 & 66.7 \\
\rowcolor{ourscolor} & \sysname{} + \scorername
& \textbf{88.5} \green{\small(+5.2)} & \textbf{86.7} \green{\small(+2.5)} & \textbf{69.4} \green{\small(+7.7)} & \textbf{76.7} \green{\small(+0.9)} & \textbf{67.7} \green{\small(+1.0)} \\
\midrule
\multirow{2}{*}{\textit{\shortstack[l]{Qwen3-4B\\-Thinking-2507}}}
& STEP + \scorername
& 86.7 & 86.7 & 60.0 & 65.0 & 67.1 \\
\rowcolor{ourscolor} & \sysname{} + \scorername
& \textbf{90.0} \green{\small(+3.3)} & \textbf{87.5} \green{\small(+0.8)} & \textbf{61.7} \green{\small(+1.7)} & \textbf{65.8} \green{\small(+0.8)} & \textbf{68.2} \green{\small(+1.1)} \\
\bottomrule
\end{tabular}
}
\end{table}

%% file: figures/experiment/analysis/hyperparameter_ablation.tex
\begin{figure}[ht]
    \centering
    \includegraphics[width=\columnwidth]{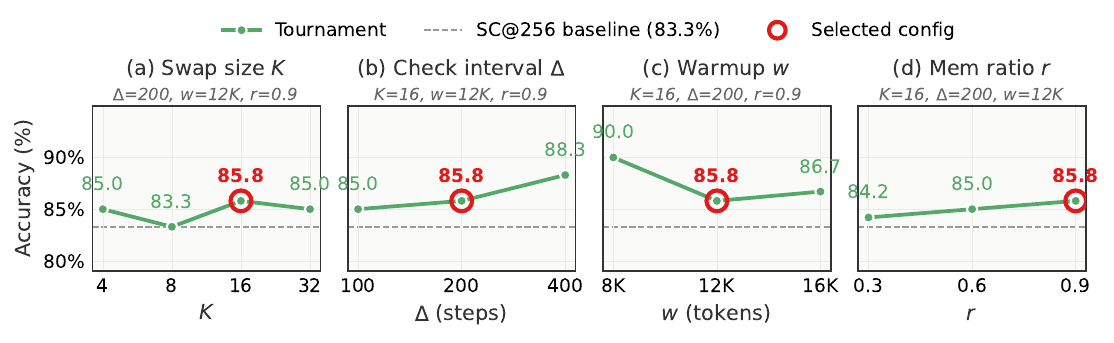}
    \vspace{-2mm}
    \caption{
    Hyperparameter sensitivity of \sysname{} on AIME-25 with DeepSeek-R1-0528-8B (batch size 256, $C{=}256$).
    Each panel varies one parameter while fixing the others to the selected configuration
    $(K{=}16, \Delta{=}200, w{=}12\mathrm{K}, r{=}0.9)$.
    The dashed line denotes the SC@256 baseline (83.3\%).
    The selected configuration (red circle) lies within a broad high-performing regime,
    consistently outperforming the baseline.
    }
    \label{fig:ablation}
    \vspace{-3mm}
\end{figure}